\documentclass[11pt]{article}

\usepackage[final]{acl}
\usepackage{makecell}
\usepackage{mdframed}
 \usepackage{xcolor}
\usepackage{subcaption}
\usepackage{placeins}
\usepackage{booktabs}
\usepackage{times}
\usepackage{latexsym}
\usepackage{multirow}
\usepackage[T1]{fontenc}

\usepackage[utf8]{inputenc}

\usepackage{microtype}

\usepackage{inconsolata}
\usepackage{graphicx}
\usepackage{pifont}
\usepackage{tcolorbox}
\usepackage{enumitem}
\usepackage{float}
\newcommand{\cmark}{\ding{51}}
\newcommand{\xmark}{\ding{55}}
\title{\textsc{PerSo}: Investigating Facets of LLMs' Social Intelligence in Real-World Persuasion Scenarios}

\author{
  \textbf{Ali Zamani\textsuperscript{*1}, Soroush Esfahanian\textsuperscript{*1}, Erfan Shahabi\textsuperscript{1},} \\
  \textbf{Mohammad Taher Pilehvar\textsuperscript{2}, Azadeh Shakery\textsuperscript{1,3}, and Yadollah Yaghoobzadeh\textsuperscript{1,4}} \\
  \textsuperscript{1}School of Electrical and Computer Engineering, College of Engineering, University of Tehran, Tehran, Iran \\
  \textsuperscript{2}Cardiff University, United Kingdom \\
  \textsuperscript{3}Institute for Research in Fundamental Sciences (IPM), Tehran, Iran \\
  \textsuperscript{4}Tehran Institute for Advanced Studies, Khatam University, Tehran, Iran \\
  \texttt{\{ali.zamani.1383, soroushesfahanian, erfanshahabi\}@ut.ac.ir}
}

\usepackage{amsmath}
\usepackage{pifont}
\usepackage{pdfrender}
\newcommand{\bolder}[1]{%
  \pdfrender{TextRenderingMode=FillStroke,LineWidth=0.2pt}{\textbf{#1}}%
}
\newcommand{\bcmark}{%
  \pdfrender{TextRenderingMode=FillStroke,LineWidth=1.2pt}{\ding{51}}%
}
\usepackage{pifont}

\usepackage{tcolorbox}
\usepackage{caption} 
\tcbuselibrary{breakable} 
\usepackage{listings}
\makeatletter
\AtBeginDocument{%
  \let\PerSoOriginalNatHyper\NAT@hyper@
  \def\NAT@hyper@#1{%
    \mbox{\PerSoOriginalNatHyper{#1}}%
  }%
}
\makeatother

\begin{document}
\maketitle
\raggedbottom
\setlength{\textfloatsep}{18pt plus 2pt minus 4pt}
\setlength{\dbltextfloatsep}{18pt plus 2pt minus 4pt}
\renewcommand{\thefootnote}{*}
\footnotetext{Equal contribution, ordered randomly.}
\renewcommand{\thefootnote}{\arabic{footnote}}
\begin{abstract}
Understanding the social intelligence of large language models (LLMs) is 
increasingly important for applications involving human-AI interaction 
and multi-agent environments. Social intelligence comprises two aspects: 
cognitive intelligence, the ability to infer others' mental states, and 
behavioral intelligence, the ability to select appropriate social actions 
based on these inferences to achieve social goals. Persuasion provides a 
natural testbed in which both aspects are simultaneously at play, yet 
prior benchmarks typically evaluate them in isolation using synthetic 
scenarios. We introduce \textsc{PerSo}, a benchmark that evaluates both 
aspects on real-world persuasion dialogues from Reddit's ChangeMyView 
community, using an LLM-in-the-loop pipeline to construct the evaluation 
tasks, validated through human annotation. Our evaluation of diverse LLMs 
shows that performance generally scales with model size but varies 
across dimensions. The strongest LLM slightly surpasses the best observed 
human performance on the behavioral task, while the best human 
participant remains ahead on the cognitive task.
\end{abstract}

\begin{table*}[!t]
\centering
\small
\setlength{\tabcolsep}{4pt}
\renewcommand{\arraystretch}{1.1}
\begin{tabular*}{\textwidth}{@{\extracolsep{\fill}}lccccc@{}}
\toprule
\textbf{Benchmark} &
\makecell{\textbf{Real Human}\\\textbf{Data}} &
\textbf{Cognitive} &
\makecell{\textbf{Cog.}\\\textbf{Aspects}} &
\makecell{\textbf{Cog.}\\\textbf{Type}} &
\textbf{Behavioral} \\
\midrule
FaNToM \citep{FANToM}
    & \xmark & \cmark & B          & R  & \xmark \\
DYNToM \citep{TDTOM}
    & \xmark & \cmark & B, E, I, A & R  & \xmark \\
PersuasiveToM \citep{PersuasiveToM}
    & \xmark & \cmark & B, D, I    & R  & \xmark \\
SocialEval \citep{SocialEvalES}
    & \xmark & \xmark & --         & -- & \cmark \\
SOTOPIA \citep{SOTOPIA}
    & \xmark & \xmark & --         & -- & \cmark \\
AgentSense \citep{agentsense}
    & \xmark & \xmark & -- & -- & \cmark \\
\bolder{\textsc{PerSo} (ours)}
    & \bcmark & \bcmark
    & \bolder{B, E, I} & \bolder{F} & \bcmark \\
\bottomrule
\end{tabular*}
\caption{
Comparison of social intelligence benchmarks.
B: Belief, D: Desire, E: Emotion, I: Intention/Intent, A: Action.
R: mental-state recognition, where the target state is inferred
from observed text;
F: mental-state forecasting, where the target state is predicted
before observing the subsequent utterance that manifests it.
}
\label{tab:RelatedWorkComparison}
\end{table*}

\section{Introduction}
\label{sec:introduction}
As LLMs become increasingly integrated into social contexts, the question of whether these models possess social intelligence has emerged as a central concern in NLP research \citep{zhu-etal-2025-social}. Their deployment in direct interaction with humans and in multi-agent environments makes assessing their social intelligence a pressing research necessity \citep{li-etal-2024-social}.

Social intelligence has been defined in different ways over time. \citet{Thorndike1920IntelligenceUses} characterized it as the ability to understand and manage people and act wisely in human relationships. \citet{Marlowe1986} emphasized understanding people's feelings, thoughts, and behaviors, including one's own, and acting appropriately on that understanding in interpersonal situations. \citet{Albrecht2006SocialIntelligence} defined it in terms of getting along with others and gaining their cooperation. \citet{Goleman2007SocialIT} distinguished social awareness---understanding others' feelings and mental states---from social facility, or using that awareness to interact effectively.

Together, these definitions characterize social intelligence as understanding others' mental states and acting appropriately to achieve social goals. Following \citet{li-etal-2024-social}, we distinguish two complementary aspects: (1) \textit{cognitive intelligence}, the ability to infer others' beliefs, emotions, and intents; and (2) \textit{behavioral intelligence}, the ability to select and execute appropriate social actions based on these inferences to achieve social goals.

A growing body of research has sought to evaluate and benchmark the social intelligence of large language models \citep{FANToM,liu2024interintentinvestigatingsocialintelligence,zhou2024sotopia,PersuasiveToM,SocialEvalES}. However, existing benchmarks largely rely on synthetic or curated data rather than real-world interactions, and tend to focus on isolated aspects of social intelligence rather than capturing it in a multi-faceted manner.

To address this gap, we introduce \textsc{PerSo}, a benchmark for
evaluating the social intelligence of LLMs across both cognitive and behavioral aspects in persuasion scenarios.
Persuasion provides a natural testbed for simultaneously assessing multiple aspects of social intelligence \citep{li-etal-2024-social, Albrecht2006SocialIntelligence}. In a successful persuasion interaction, the persuader must anticipate the mental state of the counterpart---an ability that serves as a natural foundation for measuring cognitive intelligence. Furthermore, being able to judge which of two competing responses is more persuasive to that counterpart is itself a manifestation of behavioral intelligence. Appendix~\ref{sec:appendix_construct_alignment} derives the connection between these tasks and the cited definitions of social intelligence.

To ground \textsc{PerSo} in real-world interactions, we use conversations from Reddit's ChangeMyView (CMV) community, where users engage in structured persuasion exchanges across diverse topics. To assess cognitive social intelligence, we construct multiple-choice questions targeting the persuadee's beliefs, emotions, and intents, asking models to predict each state in the persuadee's subsequent reply. This evaluates the ability to infer others' mental states from the preceding interaction. To assess behavioral social intelligence, we ask models to distinguish successful from unsuccessful persuasion attempts by selecting the response that led to persuasion, as indicated by the persuadee's delta award. This evaluates the ability to select an effective social action toward the goal of changing the persuadee's view.

In summary, our contributions are as follows:
\begin{itemize}
    \item We construct the \textsc{PerSo} dataset from real-world persuasion dialogues and propose an LLM-in-the-loop pipeline for mental-state annotation and question generation.
    \item We design targeted evaluation tasks for assessing the cognitive and behavioral aspects of social intelligence, with a single dataset jointly supporting the evaluation of
    both aspects.
    \item We evaluate a diverse set of LLMs across varying scales and architectures, revealing distinct patterns in social intelligence performance.
\end{itemize}

\section{Related Work}
\label{sec:related_works}
Recent work has introduced several benchmarks for evaluating different aspects of machine social intelligence. One line of work focuses on cognitive intelligence, assessing the capacity of LLMs to infer mental states across dialogues and social contexts \cite{CommonToM, NegotiationToM, tomato}. Some studies evaluate models' ability to reason about others' beliefs in conversational settings \cite{FANToM}. Others examine how models track the evolution of beliefs, emotions, intentions across interconnected social scenarios \cite{TDTOM}. More specifically, research on persuasive dialogues investigates models' ability to understand evolving beliefs, desires, and intentions throughout persuasion interactions \cite{PersuasiveToM}. Most of these benchmarks ask models to recognize mental states from dialogue that has already been observed. In contrast, \textsc{PERSO} asks models to forecast the persuadee's mental state before seeing their following response, as shown in Table~\ref{tab:RelatedWorkComparison}. Previous work on conversational emotions has also separated recognition from forecasting and found forecasting to be more difficult when the target utterance is not available \cite{koga2024forecasting}. Other works propose methods to improve social reasoning through narrative-based Theory of Mind (ToM) tasks \cite{COKE}. More recent work also studies cognitive ability in multi-agent interactions rather than only static or two-person settings \cite{He2023HITOMAB}. 

Alongside cognitive evaluation, other benchmarks assess behavioral intelligence through models' ability to select appropriate social actions and achieve social goals. Prior work evaluates agents in interactive role-playing scenarios \cite{SOTOPIA}, manually constructed narratives assessing goal achievement and interpersonal abilities \cite{SocialEvalES}, and individual or group level social interactions and multi-agent simulations \cite{socialBench,STSS}. More recent work examines social goal completion in multi-turn interactions alongside the inference of other agents' private information \cite{agentsense}. Unlike these primarily constructed or simulated settings, \textsc{PerSo} uses real human persuasion interactions to evaluate both cognitive and behavioral intelligence on the same data.


\begin{figure*}[t]
    \centering
    \includegraphics[width=\textwidth]{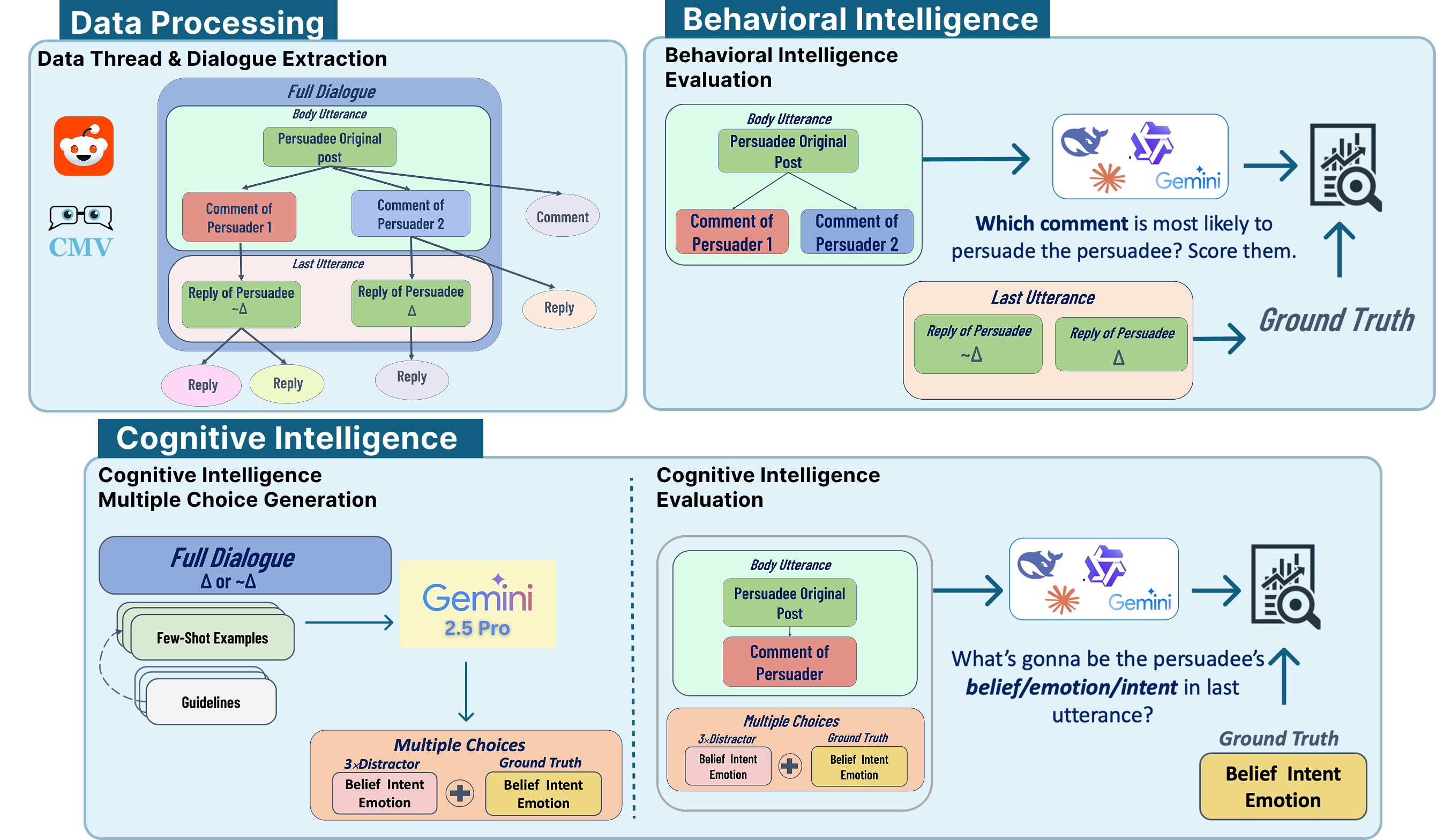}
    \caption{Overview of the \textsc{PerSo} pipeline: data extraction and 
    processing from CMV, multiple-choice question generation for 
    cognitive intelligence, and the behavioral and cognitive intelligence 
    evaluation tasks.}
    \label{fig:placeholder}
\end{figure*}
\section{Method}
\label{sec:method}

\subsection{Data Collection}
\label{sec:data_collection}

We construct our dataset using discussion threads from the Reddit CMV 
subreddit, spanning January 2024 to October 2025. CMV's structured 
environment for argumentation makes it an ideal source for naturalistic 
persuasion dialogues. In CMV, a delta ($\Delta$) is awarded when a user 
acknowledges that a comment changed their view. Reddit users can also 
upvote comments as a positive vote, and the resulting comment score 
provides a community-level signal of approval.

As shown in Figure~\ref{fig:placeholder}, we isolate threads where the
original poster (OP) presents a viewpoint and select two distinct
persuaders: one who successfully persuades the OP and one who does
not, noted with ($+$) and ($-$), respectively. For thread $i$, let $u_i^{O}$ denote the OP's initial post,
$u_{i,j}^{P}$ the response of persuader $j$, and $u_{i,j}^{O'}$ the OP's
subsequent reply, where $j \in \{+,-\}$. We represent each branch as:

\begin{equation}
D_{i,j} = \left(u_i^{O}, u_{i,j}^{P}, u_{i,j}^{O'}\right)
\end{equation}

We require that both persuaders elicit a direct reply from the OP, but 
with contrasting outcomes: in $D_{i,+}$, the OP awards a delta to the 
persuader, while in $D_{i,-}$ no delta is awarded. The $+$ and $-$ 
notations are used only to describe the dataset and are not provided to 
the evaluated models.

Because susceptibility to persuasion is subjective, we introduce an 
additional constraint to align individual outcomes with broader community 
consensus: the successful, $\Delta$-awarded comment must also possess a 
higher upvote score than its unsuccessful counterpart. This ensures the 
successful argument is socially validated. Finally, to support the 
cognitive evaluation task (Section~\ref{sec:cognitive_task}), we enforce a 
minimum length threshold of 100 tokens for both OP replies to guarantee 
sufficiently rich textual context. Following the application of these 
filtering criteria, our collection resulted in 872 qualifying threads, 
from which we randomly sampled 200 threads to form the final evaluation 
dataset.

\begin{table}[t]
\centering
\small
\setlength{\tabcolsep}{8pt}
\renewcommand{\arraystretch}{1.15}
\begin{tabular}{lrr}
\toprule
\multicolumn{3}{l}{\textbf{Dataset overview}} \\
\midrule
Threads                  & \multicolumn{2}{r}{200} \\
Date range               & \multicolumn{2}{r}{Jan 2024--Oct 2025} \\
Label balance (A\,/\,B)   & \multicolumn{2}{r}{100\,/\,100} \\

\addlinespace
\textbf{Text length (words)} & \textbf{Mean} & \textbf{Median} \\
\midrule
OP initial post          & 449 & 400 \\
Persuader response ($+$) & 242 & 182 \\
Persuader response ($-$) & 155 & 112 \\

\addlinespace
\textbf{Upvote score}     & \textbf{Mean} & \textbf{Median} \\
\midrule
Persuader response ($+$) & 50.5 & 14 \\
Persuader response ($-$) &  6.6 &  2 \\
Score margin             & 43.9 & 11 \\

\addlinespace
\multicolumn{3}{l}{\textbf{Cognitive task}} \\
\midrule
Questions per dimension  & \multicolumn{2}{r}{400 ($200 \times 2$ branches)} \\
Total MCQ instances      & \multicolumn{2}{r}{1,200} \\
\bottomrule
\end{tabular}
\caption{Summary statistics of the \textsc{PerSo} dataset.}
\label{tab:dataset_stats}
\end{table}

Table~\ref{tab:dataset_stats} summarizes key statistics of the
\textsc{PerSo} dataset. Persuasive replies are on average 56\% longer than their non-persuasive counterparts,
reflecting the natural tendency for more elaborate arguments to be
more convincing. The dataset spans diverse discussion topics,
with Ethics \&\ Morality (30\%), Politics \&\ Government (22.5\%),
and Society \&\ Culture (17\%) being the most prevalent categories, as shown in Table~\ref{tab:topic_distribution} in
Appendix~\ref{sec:appendix_topics}.

To measure the cognitive social intelligence of the evaluated models, we 
design a multiple-choice question (MCQ) task to assess their proficiency 
in inferring mental states. For each discussion branch, we formulate three 
distinct questions targeting the three core dimensions of the persuadee's 
mental state: belief, emotion, and intent \cite{li-etal-2024-social}. We implement an LLM-in-the-loop pipeline utilizing Gemini 2.5 Pro 
\cite{comanici2025gemini25pushingfrontier} to generate these questions 
through a two-stage process guided by the multiple-choice generation 
framework established by \citet{MOMENTSAC}. First, we prompt the model with 
the full dialogue $D_{i,j}$ alongside few-shot exemplars. Based on this 
context, the model identifies the OP's belief, emotion, and intent from 
$u_{i,j}^{O'}$, i.e., the OP's subsequent reply to persuader
$j$, to serve as the gold-standard choices. Second, we provide 
the model with the full dialogue and the extracted gold choices, 
instructing it to generate plausible distractors that represent 
incorrect mental states distinct from the OP's actual state. The complete prompt templates are provided in
Appendix~\ref{sec:appendix_Prompt},
Figures~\ref{fig:gold_label_prompt} and~\ref{fig:distractors_prompt}. Finally, to validate the quality of the dataset, three computer science
annotators manually reviewed a random 10\% subset of the generated
questions. We describe the annotation protocol and pairwise agreement in
Section~\ref{sec:human_annotation}, and report additional agreement
analyses and answer quality assessments in
Appendix~\ref{sec:appendix_human_annotation}. A complete sample persuasion thread is provided in Appendix~\ref{app:sample_branch77}, Figure~\ref{fig:processed_thread}.

\subsection{Task Definition}
\label{sec:task_definition}
To evaluate the social intelligence of LLMs in persuasion scenarios, we 
design two distinct tasks, each targeting one of the aspects under consideration. The first task assesses the model's cognitive intelligence through its ability to infer the mental state of the persuadee. The second task assesses the model's behavioral intelligence through its ability to identify the more persuasive comment between two alternatives. Both tasks are constructed on the same dataset collected from CMV, enabling an integrated assessment of both aspects on identical real-world data. The conceptual grounding of both tasks is detailed in Appendix~\ref{sec:appendix_construct_alignment}.
The complete prompt templates are provided in
Appendix~\ref{sec:appendix_Prompt}, 
Figures~\ref{fig:cognitive_prompt} and~\ref{fig:behavioral_prompt}. Processed examples for each task are also provided in Appendix~\ref{app:sample_branch77}, Figure~\ref{fig:processed_thread}.

\subsubsection{Behavioral Intelligence Task}
\label{sec:behavioral_task}
As shown in Figure~\ref{fig:placeholder}, to evaluate behavioral intelligence—specifically the capacity to identify and select actions that successfully achieve social goals—we use the two persuasion 
branches defined in Section~\ref{sec:data_collection}. For each thread 
$i$, the model receives

\begin{equation}
B_i = \left(u_i^{O}, u_{i,+}^{P}, u_{i,-}^{P}\right),
\end{equation}

consisting of the persuadee's original post and the two candidate 
responses from distinct persuaders. The OP's subsequent replies 
$u_{i,+}^{O'}$ and $u_{i,-}^{O'}$ are not provided. The model must 
determine which candidate response is more effective for achieving the 
social goal of persuasion.
To mitigate position bias during selection, we ask the model to rate the persuasiveness of each response on a four-point scale: \textit{not persuasive} (1), \textit{slightly persuasive} (2), \textit{persuasive} (3), and \textit{very persuasive} (4). We require the two scores to differ and take the higher-scoring response as the model's selection of the more effective social action.

The ground truth for behavioral success is determined by the delta ($\Delta$) tag awarded by the persuadee to the successful persuader in a subsequent utterance excluded from the model's input.

\begin{table*}[t]
\centering
\small
\setlength{\tabcolsep}{5pt}
\renewcommand{\arraystretch}{1.2}
\begin{tabular*}{\textwidth}{@{\extracolsep{\fill}}lcccc@{\hspace{12pt}}ccc@{}}
\toprule
\multirow{3}{*}{\textbf{Participant}}
& \multicolumn{4}{c}{\textbf{\textsc{PerSo} accuracy (\%)}}
& \multicolumn{3}{c}{\textbf{TSIS scores (\%)}} \\
\cmidrule(lr){2-5} \cmidrule(l){6-8}
& \multirow{2}{*}{Behavioral}
& \multicolumn{3}{c}{Cognitive}
& \multirow{2}{*}{\makecell{Social Information\\Processing}}
& \multirow{2}{*}{\makecell{Social\\Skills}}
& \multirow{2}{*}{\makecell{Social\\Awareness}} \\
\cmidrule(lr){3-5}
& & Belief & Emotion & Intent & & & \\
\midrule
A & 65.0 & \textbf{85.0} & \textbf{65.0} & \textbf{85.0}
  & \textbf{79} & 61 & \textbf{71} \\
B & 40.0 & 50.0 & 52.5 & 60.0 & 54 & 54 & 50 \\
C & \textbf{75.0} & 75.0 & 62.5 & 72.5
  & 50 & \textbf{75} & 64 \\
\bottomrule
\end{tabular*}
\caption{Participant-level \textsc{PerSo} accuracy and TSIS scores.
\textsc{PerSo} results were obtained before gold-label annotation on
20 behavioral instances and 40 cognitive instances per dimension,
following the protocol in Section~\ref{sec:human_setup}. TSIS is a
self-report measure \citep{Silvera2001Tromso}.
\textbf{Bold} indicates the highest value in each score column.}
\label{tab:human_results}
\end{table*}

\begin{table}[t]
\centering
\footnotesize
\setlength{\tabcolsep}{3pt}
\renewcommand{\arraystretch}{1.0}
\begin{tabular}{@{}lcccc@{}}
\toprule
Dimension & A--B & A--C & B--C & Avg. $\kappa$ \\
\midrule
Belief
& \makecell{100.0\%\\1.000}
& \makecell{90.0\%\\0.866}
& \makecell{90.0\%\\0.866}
& 0.911 \\
Emotion
& \makecell{70.0\%\\0.601}
& \makecell{70.0\%\\0.602}
& \makecell{75.0\%\\0.667}
& 0.623 \\
Intent
& \makecell{82.5\%\\0.764}
& \makecell{72.5\%\\0.630}
& \makecell{80.0\%\\0.733}
& 0.709 \\
\bottomrule
\end{tabular}
\caption{Pairwise agreement (\%, first line) and Cohen's $\kappa$
(second line) for cognitive answer selection with the final OP reply
visible.}
\label{tab:annotation_agreement}
\end{table}

\subsubsection{Cognitive Intelligence Task}
\label{sec:cognitive_task}

To evaluate the cognitive intelligence of LLMs, as shown in 
Figure~\ref{fig:placeholder}, we formulate the task as mental-state 
forecasting. 

For each dialogue branch $D_{i,j}$, the model observes:

\begin{equation}
C_{i,j} = \left(u_i^{O}, u_{i,j}^{P}\right),
\end{equation}

where $u_i^{O}$ is the persuadee's original post and $u_{i,j}^{P}$ is the 
corresponding persuader's response. The persuadee's subsequent reply 
$u_{i,j}^{O'}$ is not revealed to the model.

The model must forecast the persuadee's resulting mental state:

\begin{equation}
{M}_{i,j} = (b_{i,j}, e_{i,j}, i_{i,j}),
\end{equation}

corresponding to belief, emotion, and intent. Each dimension is evaluated 
independently using a four-option multiple-choice question containing one 
gold answer and three distractors, as discussed in Section \ref{sec:data_collection}. The model selects the option that best 
predicts the persuadee's state after receiving the persuasive response.

Importantly, although $u_{i,j}^{O'}$ is used during dataset construction 
to derive the gold mental state, it is hidden during evaluation. Therefore, 
unlike mental-state recognition from an observed utterance, the task 
requires forecasting the persuadee's state, before the utterance that 
expresses it is observed.

\subsection{Data Validation}
\label{sec:human_annotation}
To validate the reliability of the constructed benchmark, we conducted a
human annotation study on a randomly sampled subset of 10\% of the dataset:
40 instances, drawn from 200 threads containing 400 dialogues in total.
Three independent annotators, denoted as A, B, and C, evaluated the same
instances. All three were computer science students: one undergraduate
and two graduate (Master's level) students, all with prior experience
in natural language processing. Unlike the models, which must infer the persuadee's mental
state without access to the final utterance, human annotators were shown
the third utterance when making their judgments, allowing them to verify
the gold answer and distractor options against the ground-truth text. For
each dialogue, annotators were asked to identify the correct option among
the four candidates for each cognitive dimension.

In addition to answer selection, annotators evaluated the quality of the
constructed options using two separate criteria. First, they rated how
accurately the gold answer described the final utterance on a 5-point
Likert scale, ranging from 1 (\textit{completely wrong}) to 5
(\textit{completely correct}). Second, they rated the three distractor
options based on their plausibility while remaining incorrect, using a
scale from 1 (\textit{bad distractors}) to 5 (\textit{good distractors}).
To measure the reliability of the human labels, we calculated pairwise
agreement and Cohen's $\kappa$ between annotators for the answer selection
task. As shown in Table~\ref{tab:annotation_agreement}, the average
pairwise Cohen's $\kappa$ values were 0.911, 0.623, and 0.709 for belief,
emotion, and intent, respectively. These results indicate substantial
agreement among annotators, particularly for belief and intent, and
demonstrate that the human-annotated labels provide reliable supervision
for evaluating cognitive reasoning capabilities of LLMs.
Additional agreement analyses and human assessments of gold answer
correctness and distractor plausibility are reported in
Appendix~\ref{sec:appendix_human_annotation}.

To further examine the alignment of \textsc{PerSo} with an established
measure of social intelligence, all three participants completed the
Tromsø Social Intelligence Scale (TSIS) \citep{Silvera2001Tromso}, a
21-item self-report instrument comprising Social Information Processing,
Social Skills, and Social Awareness. Social Information Processing
assesses perceived ability to understand others' feelings and anticipate
their behavior, while Social Skills assesses perceived competence in
navigating social interactions. These subscales align conceptually with
our cognitive and behavioral tasks, respectively.
Table~\ref{tab:human_results} reports their TSIS scores alongside
\textsc{PerSo} accuracies obtained before gold-label annotation using the
human evaluation protocol in Section~\ref{sec:human_setup}. The behavioral
ranking is identical across \textsc{PerSo} and TSIS Social Skills
($C > A > B$). Participant A also ranks highest on Social Information
Processing and all three cognitive dimensions, although the ordering of
B and C differs between the two measures. These findings provide
preliminary evidence of alignment with the targeted aspects of social
intelligence within this small participant group.

\section{Evaluation}
\label{sec:evaluation}
\subsection{Evaluation Setup}
\label{sec:evaluation_setup}

We evaluate LLMs and human participants on the behavioral and cognitive
tasks defined in Section~\ref{sec:task_definition}, using accuracy to
measure the selection of successful persuasion responses and the
inference of persuadees' mental states, respectively. The following
subsections describe the model evaluation settings and the human
evaluation protocol.

\subsubsection{Model Evaluation}
\label{sec:model_setup}
For a comprehensive assessment of social intelligence in large language 
models, we select a diverse set of models spanning different model 
families and scales. Our evaluation set includes open-source models from 
the Gemma \citep{gemmateam2025gemma3technicalreport}, Qwen 
\citep{qwen35blog}, and DeepSeek \citep{deepseekai2025deepseekv3technicalreport} 
families at varying sizes, as well as closed-source models including 
Claude \citep{anthropic2025haiku45, anthropic2026sonnet46} and Gemini 
\citep{comanici2025gemini25pushingfrontier}. For the Qwen models, we 
additionally evaluate both reasoning and non-reasoning (with and without 
thinking) variants to examine the effect of this capability on social 
intelligence. Since the cognitive intelligence task was 
generated using Gemini 2.5 Pro, we additionally evaluate Gemini 2.5 Pro 
itself on the generated tasks to examine the potential for self-bias. To 
ensure deterministic outputs, we set the temperature to 0 and top-k to 1 
across all experiments.
We also evaluate non-LLM baselines to situate model performance relative
to surface-level cues; details of these baselines are provided in
Appendix~\ref{sec:appendix_nonllm}.

\subsubsection{Human Evaluation}
\label{sec:human_setup}
To establish a human reference point for comparison with model
performance, we administered the same tasks to three human participants
(A, B, and C), the same individuals who served as annotators in
Section~\ref{sec:human_annotation}. This evaluation
was conducted prior to the gold-label annotation stage described in
Section~\ref{sec:human_annotation}. For the
behavioral intelligence task, 20 instances (10\% of the full dataset)
were randomly sampled and given to all three participants. For the cognitive intelligence task, 40 instances (10\% of the full dataset) were randomly sampled. The results for each participant are reported in Table~\ref{tab:human_results}.

\begin{table*}[t]
	\centering
	\small
	\renewcommand{\arraystretch}{1.3}
	\setlength{\tabcolsep}{10pt}
	\begin{tabular}{lcccc}
		\hline
		\multirow{2}{*}{Model} & \multicolumn{1}{c}{Behavioral Intelligence} & \multicolumn{3}{c}{Cognitive Intelligence} \\
		\cline{2-2} \cline{3-5}
		& Behavioral & Belief & Emotion & Intent \\
		\hline
		Gemma-4B & 60.5 & 50.0 & 20.5 & 54.5 \\
		Gemma-12B & 62.5 & 58.25 & 23.25 & 63.25 \\
		Gemma-27B & \underline{71.0} & \underline{60.0} & \underline{25.75} & \underline{64.5} \\
		\hline
		Qwen-4B (W/O Thinking) & 64.0 & 66.0 & 25.25 & 60.75 \\
		Qwen-35B (W/O Thinking) & 75.0 & 71.75 & 35.75 & 73.75 \\
		Qwen-397B (W/O Thinking) & \textbf{\underline{76.0}} & \underline{80.5} & \underline{41.0} & \underline{75.0} \\
		\hline
		Qwen-4B (With Thinking) & 66.5 & 63.5 & 25.0 & 60.5 \\
		Qwen-35B (With Thinking) & \underline{73.0} & 69.75 & 41.25 & 66.5 \\
		Qwen-397B (With Thinking) & \underline{73.0} & \underline{79.0} & \underline{48.75} & \underline{74.5} \\
		\hline
		DeepSeek-V3.1 & 67.5 & 76.0 & 35.75 & 72.0 \\
		DeepSeek-V3.2 & \underline{69.0} & 73.75 & 36.0 & 69.75 \\
		DeepSeek-V4-Pro & 62.0 & \underline{80.0} & \underline{39.75} & \underline{75.0} \\
		\hline
		Haiku-4.5 & 69.5 & 68.5 & 34.5 & 72.5 \\
		Sonnet-4.6 & \underline{71.5} & \textbf{\underline{83.0}} & \textbf{\underline{59.0}} & \textbf{\underline{80.5}} \\
		\hline
		Gemini-2.5-Pro & 71.0 & 79.5 & 37.25 & 77.75 \\
		\hline
		Human (avg) & 60.0 & 70.0 & 60.0 & 72.5 \\
		Human (best) & 75.0 & 85.0 & 65.0 & 85.0 \\
		\hline
	\end{tabular}
	\caption{Behavioral and cognitive task accuracy (\%).
\textbf{Bold} marks the best model score in each column;
\underline{underlining} marks the best score within each multi-row model
block, including ties, with Qwen thinking modes treated separately.
Human (avg/best) reports the mean/columnwise maximum across three
participants.
}
	\label{tab:results}
\end{table*}

\subsection{Behavioral Intelligence Results}
\label{sec:behavioral_results}
Table~\ref{tab:results} reveals several notable patterns in model 
performance on the behavioral intelligence task. The most apparent 
pattern is a positive relationship between model scale and performance 
within most model families. In the Gemma family, the 12B variant 
achieves slightly higher accuracy than the 4B variant at 60.5\%, and 
this accuracy further increases to 71\% for the 27B variant. A similar 
overall pattern is observed in the Qwen family. In the without-thinking 
setting, model performance improves consistently with model scale, while 
in the with-thinking setting, performance increases from the 4B to the 
35B variant and then remains unchanged for the 397B variant.

Contrary to expectation, this pattern breaks down within the DeepSeek 
family. The newest variant, V4-Pro, exhibits the weakest performance 
among the three variants in this family at 62\%, compared to 67.5\% for 
V3.1 and 69\% for V3.2. This discrepancy may stem from differences in 
training data or the fine-tuning process specific to this variant. It 
should be noted that DeepSeek models were evaluated without the reasoning 
capability to ensure a fair comparison with other models.

Another notable observation is the close competition between the 
best-performing open-source and closed-source models. Qwen-3.5-397B, as 
an open-source model, achieves the highest performance on this task at 
76\%, while the best-performing closed-source model, Sonnet-4.6, 
reaches 71.5\%, followed closely by Gemini-2.5-Pro at 71.0\%. This finding runs 
counter to the common assumption that closed-source models consistently 
outperform open-source ones on complex social tasks, and suggests that 
sufficient scale in an open-source model can close, or even reverse, 
this gap.

The best-performing model, Qwen-3.5-397B without thinking, achieves 76\% accuracy, only 1 percentage point above the best human participant at 75\%. This small difference indicates that the strongest model and the strongest human perform at a comparable level on this task.

To examine the robustness of these results to reply length, we repeat 
the evaluation after removing outlier instances with an extreme length 
difference between the two replies (length ratio above the third 
quartile). On the resulting subset of 167 instances, all models show a 
modest decrease in accuracy, while the overall pattern of results is 
preserved: the relative ordering of models remains largely unchanged, 
and Qwen-3.5-397B remains the best-performing model. The complete 
results are provided in Appendix~\ref{sec:appendix_length_control}, and 
comparisons with non-LLM baselines in Appendix~\ref{sec:appendix_nonllm}.

\subsection{Cognitive Intelligence Results}
\label{sec:cognitive_results}
Table~\ref{tab:results} reveals several distinct patterns across the 
three cognitive dimensions. Most notably, \textit{emotion} is 
consistently the most difficult dimension for all models. On a 
four-option question, chance-level accuracy is 25\%, yet Gemma-3-4B and 
Gemma-3-12B reach only 20.5\% and 23.25\% respectively on this 
dimension, at or below chance, while the same models perform 
considerably better on belief and intent. This aligns with the human 
annotation data, where inter-annotator agreement was also lowest for 
emotion, suggesting that inferring emotional states is inherently harder 
than inferring beliefs or intents, for both models and humans alike.

Another interesting observation concerns the spread of performance across 
dimensions. Emotion shows the widest variation, ranging from 20.5\% to 
59\%, while most models cluster within a narrower 60--80.5\% band on 
intent. This suggests that intent identification is less sensitive to 
a model's specific capabilities and tends to be achieved at a relatively 
consistent level across different models.

The positive scale--performance relationship seen in the behavioral task 
also holds here; for instance, within the Qwen family, belief accuracy 
rises from 66\% for the 4B variant to 80.5\% for the 397B variant. 
Interestingly, the DeepSeek family, which showed an irregular trend in 
the behavioral task, does not exhibit the same issue here. V4-Pro 
achieves the highest scores within the DeepSeek family on all three aspects. This contrast suggests that V4-Pro's weaker 
behavioral performance is likely tied to a specific limitation in 
comparing two competing arguments, rather than a general weakness in 
understanding mental states.

Finally, Sonnet-4.6 stands out on the emotion dimension with a score of 
59\%, more than 10 percentage points above the second-best model on this 
dimension, Qwen-3.5-397B with thinking at 48.75\%. This gap is not 
observed on either of the other two dimensions and may point to a 
stronger capacity in this model for detecting subtle emotional states.

It is also worth noting that Gemini-2.5-Pro does not achieve the
highest score on any of the three cognitive dimensions. It reaches
79.5\% on belief, 37.25\% on emotion, and 77.75\% on intent, trailing
Sonnet-4.6 on all three dimensions. It also trails Qwen-3.5-397B with
thinking on emotion. Given that Gemini-2.5-Pro was involved in generating
the cognitive task itself, its lack of consistent dominance provides
evidence against systematic bias favoring this model in the evaluation.

A comparison with the best human participant reveals relatively small
gaps across the three cognitive dimensions. Sonnet-4.6, the strongest
model on all three dimensions, scores 2 percentage points lower than the
best human participant on belief, 6 points lower on emotion, and 4.5
points lower on intent. Thus, the best observed human performance remains
higher across all three cognitive dimensions, although the differences
are relatively small.

Further comparisons with non-LLM-based models are provided in
Appendix~\ref{sec:appendix_nonllm}.

\subsection{Thinking Analysis}
\label{sec:thinking_analysis}
To examine the effect of the thinking capability, we compared different 
variants of Qwen-3.5 with and without this capability enabled. 
Table~\ref{tab:results} shows that thinking has different effects across 
dimensions, and no single conclusion can be drawn for all of them.

The clearest pattern appears on the emotion dimension. Enabling thinking 
improves accuracy for the 35B and 397B variants; the 35B variant goes 
from 35.75\% to 41.25\%, and the 397B variant from 41.0\% to 48.75\%. 
The 4B variant, however, remains almost unchanged, from 25.25\% to 
25.0\%. Belief tells a different story: thinking lowers performance 
across all three sizes, with the 4B variant dropping from 66.0\% to 
63.5\%, the 35B from 71.75\% to 69.75\%, and the 397B from 80.5\% to 
79.0\%.

This difference likely comes down to the nature of each dimension. 
Emotion relies on subtle and implicit textual cues, and step-by-step 
reasoning can help the model interpret them more carefully. Belief, on 
the other hand, tends to be expressed more explicitly in the text, and 
thinking in this case adds complexity rather than clarity.

For intent and the behavioral task, there is no clear pattern. On intent, 
all three variants show a slight drop with thinking enabled, though the 
decline is more pronounced for the 35B variant, from 73.75\% to 66.5\%. 
On the behavioral task, the results are mixed: the 4B variant improves, 
while the 35B and 397B variants perform worse with thinking enabled. 
This inconsistency suggests that thinking offers no clear benefit for 
these two dimensions and may in some cases hurt performance.

Overall, thinking is most effective where the text is more ambiguous and 
the model needs to reason more deeply. For dimensions where the relevant 
information is stated more explicitly, enabling thinking does not help 
much.


\section{Conclusion}
\label{sec:conclusion}

In this work, we introduced \textsc{PerSo}, a benchmark for evaluating 
the social intelligence of large language models across cognitive and 
behavioral dimensions, using persuasion scenarios as the evaluation 
testbed. \textsc{PerSo} is built on real-world conversations from Reddit's 
ChangeMyView community and enables an integrated evaluation of 
both dimensions on one dataset, addressing a key limitation of prior 
benchmarks that either evaluate these dimensions in isolation or rely on 
synthetic data.

Our evaluation across a diverse set of LLMs reveals several consistent 
patterns. Model performance generally improves with scale, though not 
universally. The \textit{emotion} dimension is consistently the most 
difficult to infer, both for models and human participants, while belief, 
intent, and overall persuasiveness prove comparatively easier. Enabling 
reasoning selectively improves emotion inference but offers 
little benefit, and occasionally slight harm, on the other dimensions. Finally, the strongest model slightly surpasses the best observed human
performance on the behavioral task, while the best human participant
remains above the top-performing model on all three cognitive dimensions.
The gaps suggest that cognitive social reasoning---particularly
emotion---continues to present a meaningful challenge for
current LLMs.

\section*{Limitations}

Our human reference results are based on only three participants, 
which limits the statistical reliability of any comparison between 
human and model performance. A larger and more diverse participant 
pool would be needed to draw more definitive conclusions about the 
relative standing of humans and models on these tasks.

\textsc{PerSo} is constructed exclusively from the ChangeMyView 
subreddit, which, while a rich source of naturalistic persuasion 
dialogue, represents a specific community with its own norms, 
demographics, and argumentative conventions. Moreover, our data is 
limited to English-language conversations. The extent to which our 
findings generalize to other persuasion contexts, languages, and 
cultural settings remains an open question. Furthermore, the 
dialogues in this work are limited to single-turn exchanges between 
the persuader and the persuadee. Investigating persuasion scenarios 
with a greater number of turns and longer conversations (which may 
reveal distinct patterns of social intelligence) was not feasible 
in this work due to constraints on the volume of qualifying data, 
and represents a valuable direction for future research.

\section*{Ethical Considerations}
The human annotation and evaluation studies reported in this work
involved three volunteer participants who provided informed consent
prior to participation. Participants were made aware of the nature
of the tasks and their right to withdraw at any time. All individual
results, including scores on the Tromsø Social Intelligence Scale
(TSIS), are reported in anonymized form using labels A, B, and C,
with no personally identifiable information disclosed. The TSIS
scores reflect personal psychological assessments and are included
solely to provide context for interpreting human performance; they
should not be used to make judgments about individual participants.
The Reddit data used in this work was collected from publicly
available posts and comments on the ChangeMyView subreddit and does
not include any private communications.

\bibliography{custom}

\clearpage
\appendix

\section{Appendix}

\subsection{Conceptual Grounding of the Evaluation Tasks}
\label{sec:appendix_construct_alignment}

This section makes explicit the connection between the psychological
definitions in Section~\ref{sec:introduction} and the tasks in
Section~\ref{sec:task_definition}. The argument proceeds from the
definition of social intelligence, to the abilities it involves, to
their expression in persuasion, and finally to the observable decisions
scored by \textsc{PerSo}.

\paragraph{Social intelligence and its cognitive and behavioral facets.}
The definitions discussed in the introduction share a connection
between understanding people and acting effectively in interpersonal
situations. \citet{Thorndike1920IntelligenceUses} associates social
intelligence with understanding people and acting wisely in human
relationships. \citet{Marlowe1986} makes this connection explicit:
social intelligence involves understanding people's thoughts, feelings,
and behaviors, including one's own, and acting appropriately on that
understanding. \citet{Goleman2007SocialIT} similarly distinguishes
\emph{social awareness} from \emph{social facility}: understanding
another person and using that awareness in effective interaction.

These accounts motivate our focus on two complementary facets:
cognitive understanding and behavioral effectiveness. Furthermore, we use the
definitions proposed by \citet{li-etal-2024-social} for these two facets.

\paragraph{Cognitive intelligence: from understanding mental states
to forecasting a recipient's reaction.}
The cognitive facet concerns understanding another person's mental
states from available social cues. This follows the understanding
component of \citet{Marlowe1986} and the social-awareness component of
\citet{Goleman2007SocialIT}. More specifically,
\citet{li-etal-2024-social} organize cognitive intelligence around
beliefs, emotions, and intents. These correspond to the three targets
of our cognitive task: what the persuadee regards as true, how they
feel, and what they seek to do or communicate.

In a persuasion scenario, this understanding concerns a recipient who
already holds a particular view and is presented with a message
intended to influence it. Given the OP's initial post $u_i^{O}$ and a
persuader's response $u_{i,j}^{P}$, \textsc{PerSo} asks the model to
forecast the OP's resulting belief, emotion, or intent by selecting
among four candidate descriptions. The relevant judgment is therefore
how this particular recipient will respond to this particular message.
It is not whether the model itself endorses the argument or considers
it objectively correct.

The subsequent reply $u_{i,j}^{O'}$ supplies evidence for constructing
the reference answer but is withheld during evaluation. This separates
the evidence used to establish the target from the information
available to predict it. Consequently, the task evaluates anticipation
of a mental state expressed in a future response, rather than
extraction of that state from an already visible utterance.
Forecasting is thus the operational form that mental-state
understanding takes in our persuasion setting; it is not proposed as
an exhaustive definition of cognitive intelligence. The reference
answers remain interpretations of expressed states, supported by the
annotation study in Section~\ref{sec:human_annotation}, rather than
direct observations of private mental experience.

\paragraph{Behavioral intelligence: from effective social action
to response selection.}
The behavioral facet concerns using social understanding to communicate
and act effectively toward social goals \citep{li-etal-2024-social}.
It corresponds to acting on interpersonal understanding in
\citet{Marlowe1986} and to social facility in
\citet{Goleman2007SocialIT}. In a text-based persuasion exchange, a
response is a communicative action, and changing the recipient's view
is the local social goal. Selecting between candidate responses
therefore provides a constrained test of the action-selection
component of this facet.

Our behavioral task holds the OP and initial post fixed while
presenting two naturally occurring responses. The model assigns
unequal persuasiveness scores to the responses, thereby selecting the
action it judges more effective for this recipient. The reference
selection is grounded in the OP's observed delta award, with both
subsequent OP replies hidden from the model. An observed interpersonal
outcome is therefore used to evaluate a prospective decision: given
this recipient and these available responses, which response should
be selected to achieve the persuasion goal?

The criterion is recipient-specific persuasive success rather than an
evaluator's general preference for an argument. The additional upvote
constraint aligns the selected outcome with a community-level signal;
the delta remains the outcome label. A response without a delta lacks
this explicit success signal, but may still have affected the
recipient. Likewise, the task assesses judgment about which action
to select, rather than the ability to generate and execute a
persuasive response in live interaction.

\paragraph{Why persuasion brings the two facets together.}
The connection to persuasion is supported by the cited accounts of
social intelligence. \citet{Albrecht2006SocialIntelligence} emphasizes
gaining cooperation and describes clarity in terms of communicating
ideas effectively, including persuasion.
\citet{Goleman2007SocialIT} includes influence within social facility,
and \citet{li-etal-2024-social} place persuasion under behavioral
intelligence. Persuasion consequently supplies a natural interpersonal
setting with an identifiable recipient, a communicative action, and a
social goal. Within this setting, understanding the recipient's
reaction and selecting an effective response are complementary
questions.

The processed example in Figure~\ref{fig:processed_thread} illustrates
the distinction. After the medical-progress argument, the OP concedes
the need for further medical advances while retaining a broader
concern about well-being. The cognitive questions target this
selective belief revision, the receptive emotional stance, and the
intent to qualify the original claim. The behavioral question instead
asks which of the two responses achieved the acknowledged view
change. Thus, the tasks evaluate different aspects of the same
interaction: the recipient's resulting mental state and the
effectiveness of a communicative action.

\paragraph{From the task definitions to the measured construct.}
The reasoning chain is therefore as follows. Established definitions
connect interpersonal understanding with effective social action.
Persuasion gives these abilities a concrete recipient, context, and
goal. Mental-state forecasting operationalizes the understanding
component, while outcome-grounded response selection operationalizes
the action-selection component. Evaluating both on the same
interactions therefore provides a definition-grounded assessment of
complementary facets of social intelligence.

This establishes the conceptual relevance of the tasks without
equating their scores with complete psychometric validation of social
intelligence. The tasks need not be cognitively independent, and their
scores do not establish that a model internally uses its forecast
when selecting a response. Nor does persuasive success alone establish
social wisdom or competence in every interpersonal setting.
Human annotation, the exploratory TSIS comparison, and the baseline
analyses provide additional evidence with different purposes.
Within these boundaries, \textsc{PerSo} measures mental-state
forecasting and goal-directed response selection as theoretically
grounded cognitive and behavioral facets of social intelligence in
persuasion.

\subsection{Additional Annotation Results}
\label{sec:appendix_human_annotation}
This section reports additional analyses of the human annotation study
described in Section~\ref{sec:human_annotation}.

\paragraph{Multi-annotator agreement.}
To further validate the robustness of these agreement estimates, we also
computed Fleiss' $\kappa$ and Krippendorff's $\alpha$ across all three
annotators jointly. As shown in Table~\ref{tab:multi_metric_agreement},
both metrics closely match the average pairwise Cohen's $\kappa$ for each
dimension, confirming that the reported agreement levels are consistent
across multiple reliability measures rather than being an artifact of a
single metric.

\begin{table}[htbp]
\centering
\setlength{\tabcolsep}{6pt}
\begin{tabular}{l|cc}
\hline
Dimension & Fleiss' $\kappa$ & \makecell{Krippendorff's $\alpha$} \\
\hline
Belief  & 0.911 & 0.911 \\
\hline
Emotion & 0.622 & 0.625 \\
\hline
Intent  & 0.708 & 0.711 \\
\hline
\end{tabular}
\caption{Multi-annotator agreement across all three annotators (A, B, and
C) jointly, using two complementary reliability metrics.}
\label{tab:multi_metric_agreement}
\end{table}

\paragraph{Answer quality.}
For the quality assessment of gold answers and distractor options, we
report descriptive statistics rather than agreement coefficients. Since
most ratings were concentrated at the upper end of the scale,
agreement-based reliability measures are not informative due to the
resulting ceiling effect. The average scores demonstrate that the gold
answers accurately capture the persuadee's final mental states and that
the distractor options are generally plausible while remaining incorrect.
The detailed results are presented in Table~\ref{tab:option_quality}.

\begin{table}[htbp]
\centering
\setlength{\tabcolsep}{4pt}
\begin{tabular}{l|cc}
\hline
Dimension & \makecell{Gold\\Correctness} & \makecell{Distractor\\Plausibility} \\
\hline
Belief  & 4.775 $\pm$ 0.474 & 4.575 $\pm$ 0.628 \\
\hline
Emotion & 4.517 $\pm$ 0.866 & 4.258 $\pm$ 0.780 \\
\hline
Intent  & 4.750 $\pm$ 0.520 & 4.558 $\pm$ 0.739 \\
\hline
Overall & 4.681 $\pm$ 0.650 & 4.464 $\pm$ 0.716 \\
\hline
\end{tabular}
\caption{Human evaluation of gold answer correctness and distractor
plausibility using a 5-point Likert scale, with the third utterance
visible to annotators. Values are reported as mean $\pm$ standard
deviation.}
\label{tab:option_quality}
\end{table}
\subsection{Behavioral vs.\ Cognitive Intelligence}
\label{sec:beh_vs_cog}
A cross-task comparison reveals that, within most model families, models 
that perform better on the cognitive intelligence task also tend to 
perform better on the behavioral intelligence task, suggesting a degree 
of shared underlying capability. However, this correlation does not imply 
that the two tasks measure identical aspects of social intelligence, as 
the magnitude of improvement across model sizes within a family varies 
substantially between the two tasks.

In the Gemma family, the behavioral dimension shows a larger performance 
spread across model sizes (10.5 percentage points from 4B to 27B) 
compared to the average spread across cognitive dimensions 
(approximately 8.4 percentage points), suggesting that behavioral 
intelligence is more sensitive to scale within this family. The pattern 
reverses in the Qwen and Claude families: Qwen models without thinking 
exhibit a larger margin on cognitive dimensions (approximately 15 
percentage points on average across belief, emotion, and intent) than on 
the behavioral task (12 percentage points), while in the Claude family 
this contrast is even more pronounced: the behavioral gap between 
Haiku-4.5 and Sonnet-4.6 is only 2.0 percentage points, whereas the 
cognitive gap averages approximately 15.7 percentage points across the 
three dimensions.

The DeepSeek family presents a notably different picture: whereas V4-Pro 
is the weakest of the three variants on the behavioral task (62.0\%, 
compared to 67.5\% for V3.1 and 69.0\% for V3.2), it consistently 
achieves the highest scores among its family variants on all three 
cognitive dimensions (80.0\% on belief, 39.75\% on emotion, and 75.0\% 
on intent). This reversal in the performance ordering between the two 
tasks indicates that the behavioral and cognitive dimensions of social 
intelligence can be dissociated, and that architectural or training 
differences between model variants may selectively affect one dimension 
while leaving the other largely unaffected.

Taken together, these observations suggest that behavioral and cognitive 
social intelligence, while positively correlated at a broad level, are 
distinct capabilities that scale differently across model families and 
sizes. A high score on one dimension should therefore not be taken as a 
reliable proxy for performance on the other.

\subsection{Behavioral Answer Distribution Analysis}
\label{sec:appendix_beh_distribution}

\begin{figure*}[t]
	\centering
	\includegraphics[width=\textwidth]{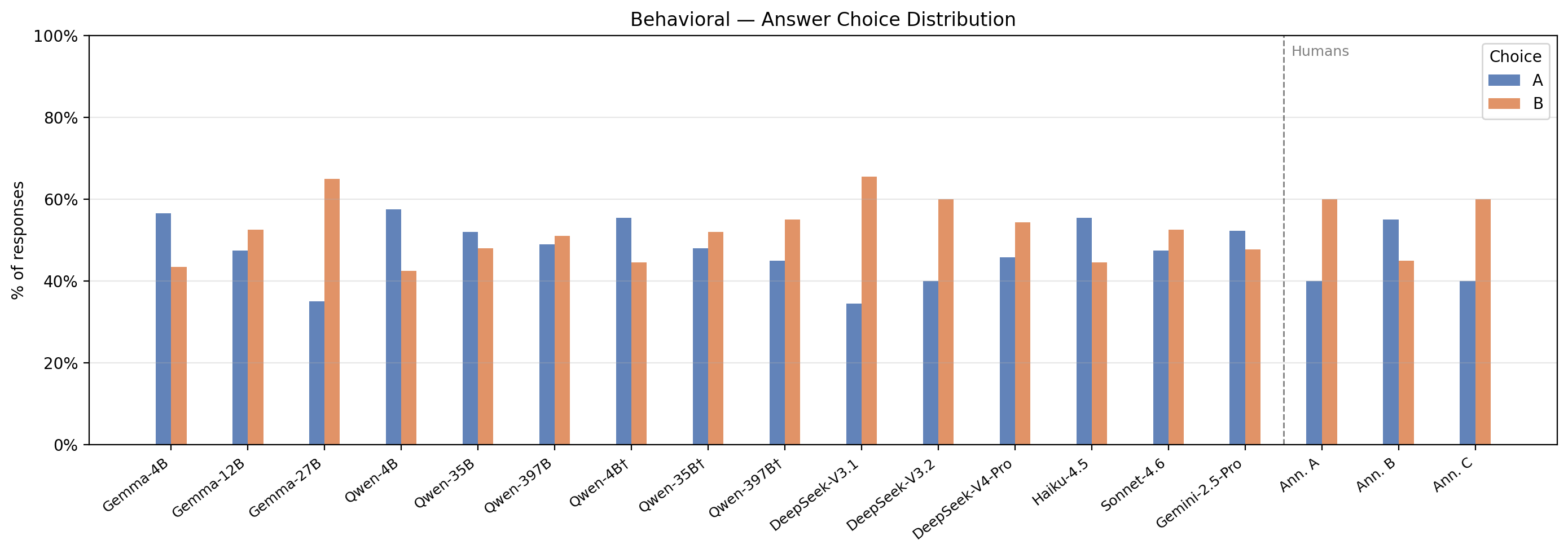}
	\caption{Distribution of choice A versus choice B on the behavioral 
		intelligence task, for each model and human annotator. The dagger 
		symbol ($\dagger$) denotes models evaluated with thinking enabled.}
	\label{fig:beh_distribution}
\end{figure*}

\begin{figure*}[t]
	\centering
	\includegraphics[width=\textwidth]{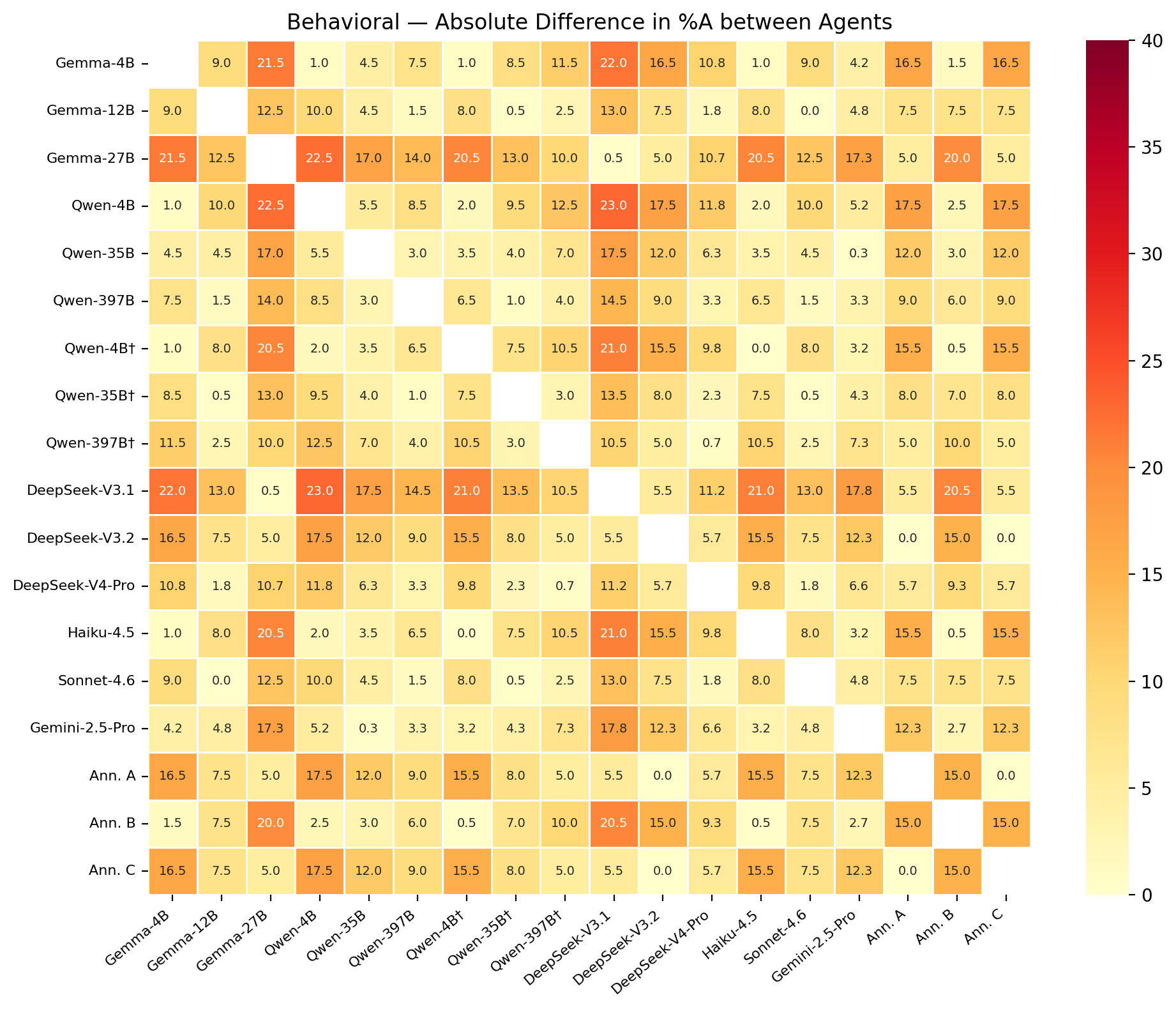}
	\caption{Absolute pairwise difference in \%A between all evaluated 
		models and human annotators on the behavioral intelligence task. The 
		dagger symbol ($\dagger$) denotes models evaluated with thinking 
		enabled. Model proportions are computed over all instances, whereas 
		human proportions are computed on the instances shown to the 
		annotators; differences involving human annotators should 
		therefore be interpreted with caution.}
	\label{fig:beh_diff_heatmap}
\end{figure*}

Figure~\ref{fig:beh_distribution} shows the distribution of choice A 
versus choice B for each model and human annotator on the behavioral 
intelligence task. Since the display order of the two persuaders in the 
prompt was randomized, this distribution effectively measures 
\textit{position bias}, a systematic tendency to favor the first or 
second option independent of its content. Most models are close to 
balanced: 12 of the 15 models select the first option between 45\% and 
58\% of the time, while Gemma-27B, DeepSeek-V3.1, and DeepSeek-V3.2 show 
a clear preference for the second option (60\% to 65.5\%). 
Figure~\ref{fig:beh_diff_heatmap} summarizes these differences as the 
absolute pairwise difference in the proportion of A choices. For the 
human annotators, the proportions are based on the instances shown to 
them; since the gold answer appeared in the second position in most of 
these instances, the human distributions partly reflect answer 
correctness rather than position bias and should be interpreted with 
caution.

\subsection{Cognitive Answer Distribution Analysis}
\label{sec:appendix_cog}

\begin{figure*}[t]
	\centering
	\includegraphics[width=\textwidth]{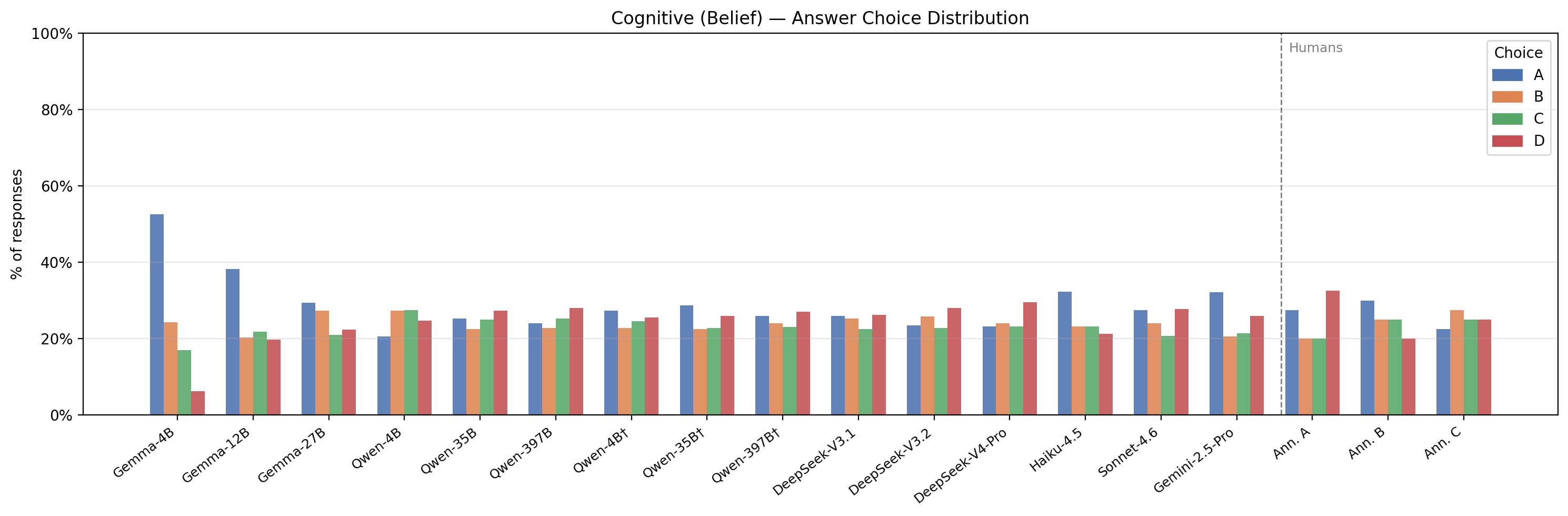}
	\caption{Distribution of choices A, B, C, and D for the belief 
		dimension of the cognitive intelligence task, for each model and 
		human annotator. The dagger symbol ($\dagger$) denotes models 
		evaluated with thinking enabled.}
	\label{fig:cog_belief_distribution}
\end{figure*}

\begin{figure*}[t]
	\centering
	\includegraphics[width=\textwidth]{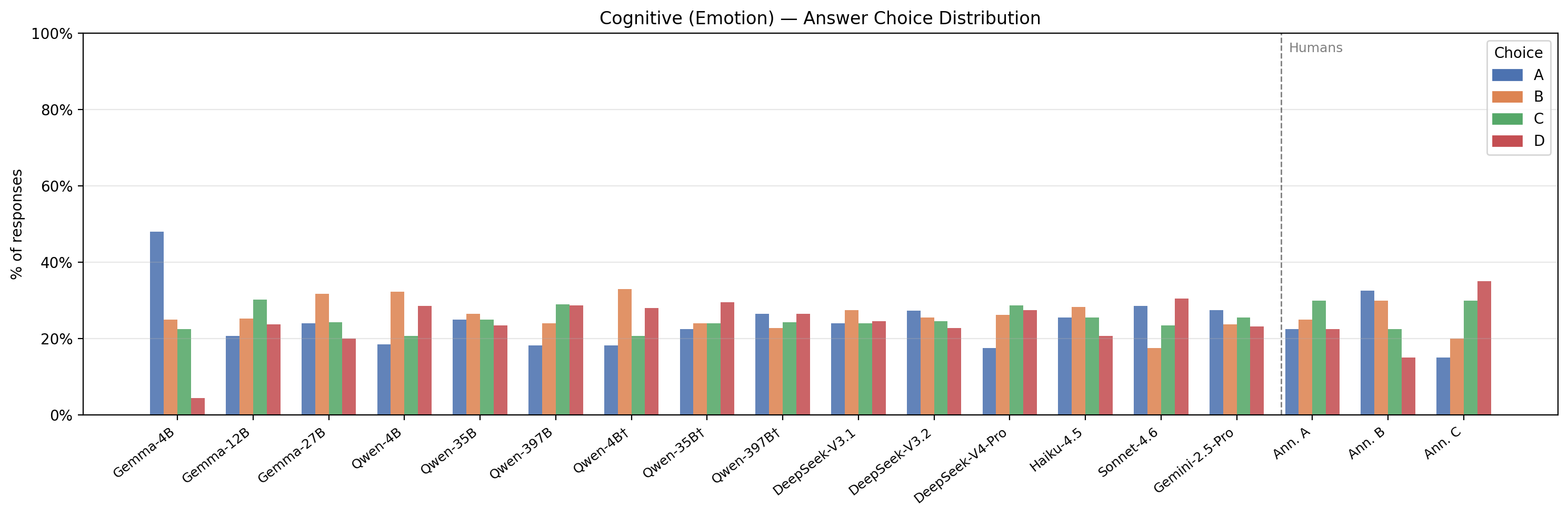}
	\caption{Distribution of choices A, B, C, and D for the emotion 
		dimension of the cognitive intelligence task, for each model and 
		human annotator. The dagger symbol ($\dagger$) denotes models 
		evaluated with thinking enabled.}
	\label{fig:cog_emotion_distribution}
\end{figure*}

\begin{figure*}[t]
	\centering
	\includegraphics[width=\textwidth]{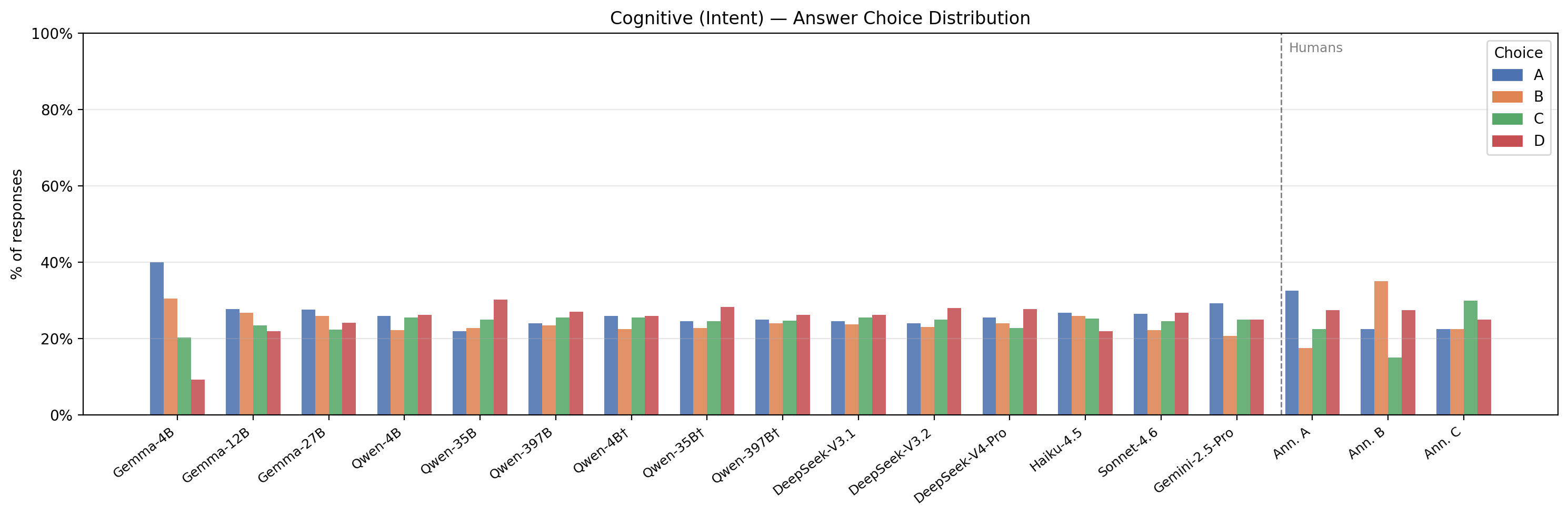}
	\caption{Distribution of choices A, B, C, and D for the intent 
		dimension of the cognitive intelligence task, for each model and 
		human annotator. The dagger symbol ($\dagger$) denotes models 
		evaluated with thinking enabled.}
	\label{fig:cog_intent_distribution}
\end{figure*}

Figures~\ref{fig:cog_belief_distribution}, 
\ref{fig:cog_emotion_distribution}, and~\ref{fig:cog_intent_distribution} 
show the distribution of choices A, B, C, and D for each cognitive 
dimension. Since one of the four options is always the gold answer and 
its position was randomized, these distributions reflect both the 
model's accuracy and its positional preference. Most models distribute 
their choices fairly evenly across the four options, with each option 
selected between 17.5\% and 33\% of the time. The main exception is 
Gemma-4B, which shows a consistent preference for option A across all 
three dimensions (52.5\% on belief, 48\% on emotion, and 40\% on intent) 
and rarely selects option D, suggesting a stable positional preference 
of this model rather than an effect of question content. A milder 
preference for option A is also observed for Gemma-12B on belief 
(38.25\%), although it does not persist on emotion or intent.

\begin{figure*}[t]
	\centering
	\includegraphics[width=\textwidth]{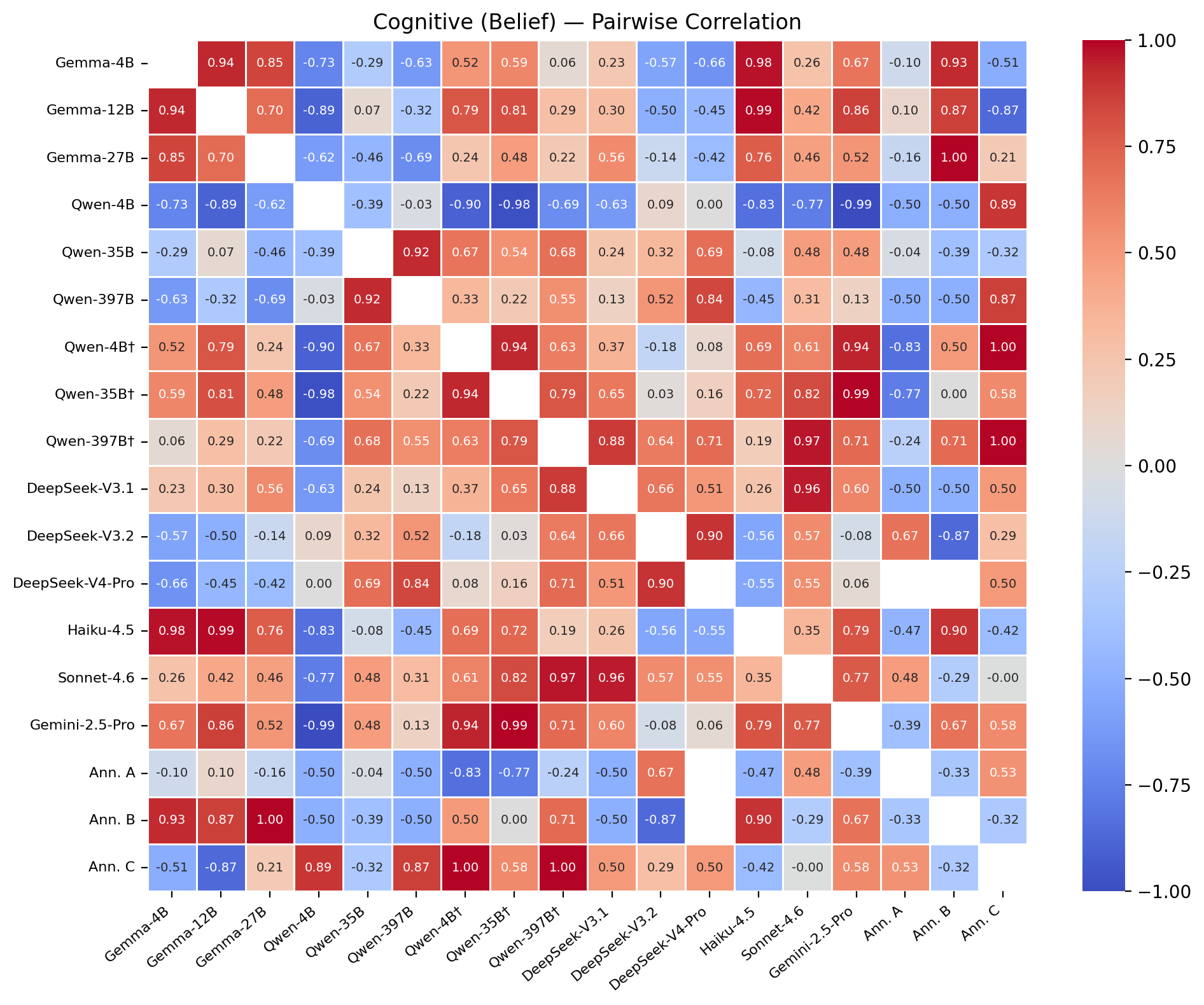}
	\caption{Pairwise Pearson correlation of answer-choice distributions 
		between all evaluated models and human annotators, for the belief 
		dimension of the cognitive intelligence task. Correlations involving 
		human annotators are computed on shared instances. Blank cells 
		indicate undefined correlations, which occur when a distribution 
		computed on the shared instances is exactly uniform. The dagger 
		symbol ($\dagger$) denotes models evaluated with thinking enabled.}
	\label{fig:cog_belief_correlation}
\end{figure*}

\begin{figure*}[t]
	\centering
	\includegraphics[width=\textwidth]{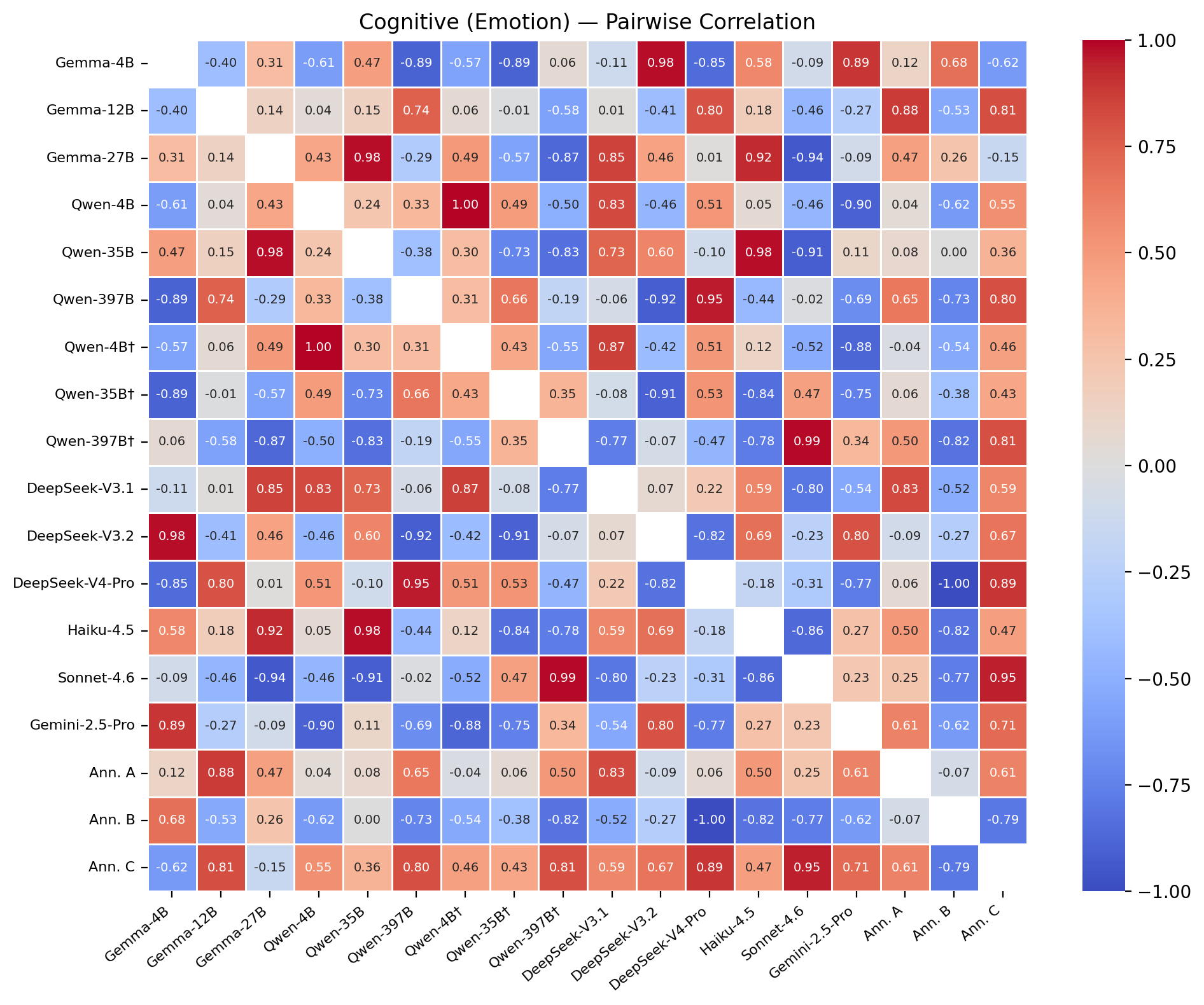}
	\caption{Pairwise Pearson correlation of answer-choice distributions 
		between all evaluated models and human annotators, for the emotion 
		dimension of the cognitive intelligence task. Correlations involving 
		human annotators are computed on shared instances. The dagger symbol 
		($\dagger$) denotes models evaluated with thinking enabled.}
	\label{fig:cog_emotion_correlation}
\end{figure*}

\begin{figure*}[t]
	\centering
	\includegraphics[width=\textwidth]{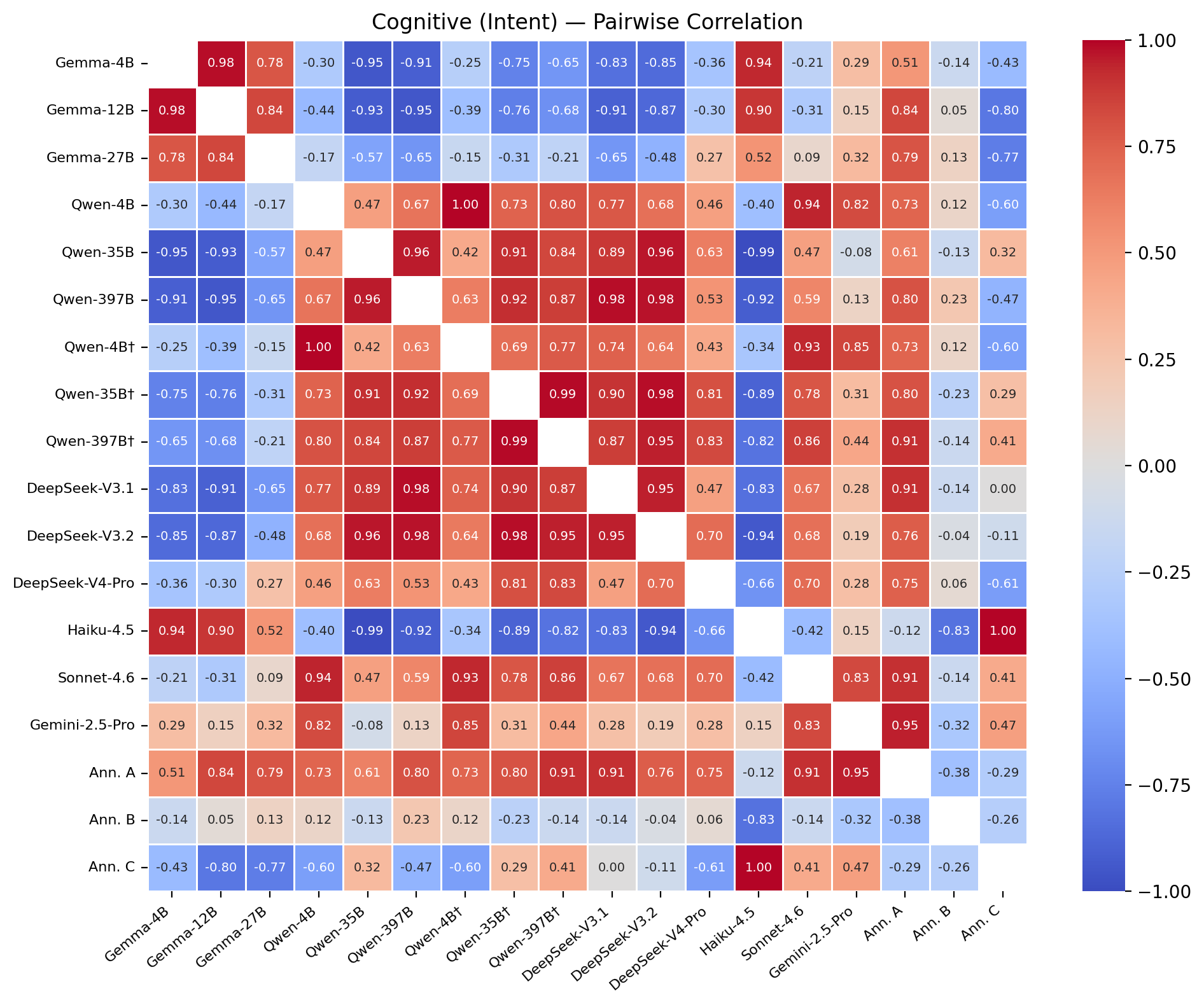}
	\caption{Pairwise Pearson correlation of answer-choice distributions 
		between all evaluated models and human annotators, for the intent 
		dimension of the cognitive intelligence task. Correlations involving 
		human annotators are computed on shared instances. The dagger symbol 
		($\dagger$) denotes models evaluated with thinking enabled.}
	\label{fig:cog_intent_correlation}
\end{figure*}

Figures~\ref{fig:cog_belief_correlation}, 
\ref{fig:cog_emotion_correlation}, and~\ref{fig:cog_intent_correlation} 
report the pairwise Pearson correlation between the choice distributions 
of all agents. Correlations between models are computed over all 
instances, whereas correlations involving a human annotator are computed 
on the instances shared by the two agents, so that both distributions 
refer to the same questions; for these pairs, the model's distribution 
is therefore recomputed on the shared instances and may differ from the 
full-set distribution shown in Figures~\ref{fig:cog_belief_distribution}, 
\ref{fig:cog_emotion_distribution}, and~\ref{fig:cog_intent_distribution}. 
Across all three dimensions, correlations between models span almost the 
full range from strongly negative to strongly positive, indicating that 
models differ considerably in their overall selection patterns. 
Correlations between human annotators and models vary in sign and 
magnitude across annotators and dimensions; for example, Annotator A 
correlates positively with nearly all models on intent, whereas 
Annotator B correlates negatively with most models on emotion. 
Correlations among the annotators themselves are low or negative in 
most cases. Since each correlation is computed over only four 
proportions, these results describe the similarity of overall selection 
patterns rather than agreement on individual questions.


\clearpage

\subsection{Topic Distribution}
\label{sec:appendix_topics}

Table~\ref{tab:topic_distribution} shows the distribution of
discussion topics across the 200 threads in \textsc{PerSo}.
The dataset covers a broad range of domains, with Ethics
\&\ Morality, Politics \&\ Government, and Society \&\ Culture
accounting for the majority of threads.

\begin{table}[h]
\centering
\small
\setlength{\tabcolsep}{10pt}
\renewcommand{\arraystretch}{1.15}
\begin{tabular}{lrr}
\toprule
\textbf{Topic} & \textbf{Count} & \textbf{\%} \\
\midrule
Ethics \&\ Morality     & 60 & 30.0 \\
Politics \&\ Government & 45 & 22.5 \\
Society \&\ Culture     & 34 & 17.0 \\
Philosophy \&\ Religion & 30 & 15.0 \\
Economics \&\ Business  &  9 &  4.5 \\
Science \&\ Technology  &  7 &  3.5 \\
International Affairs   &  7 &  3.5 \\
Education \&\ Academia  &  6 &  3.0 \\
Health \&\ Medicine     &  2 &  1.0 \\
\midrule
\textbf{Total} & \textbf{200} & \textbf{100.0} \\
\bottomrule
\end{tabular}
\caption{Topic distribution of the 200 CMV threads in
\textsc{PerSo}.}
\label{tab:topic_distribution}
\end{table}

\subsection{Non-LLM Baseline Results}
\label{sec:appendix_nonllm}

\begin{figure*}[t]
\centering
\begin{minipage}{0.57\textwidth}
    \centering
    \includegraphics[width=\linewidth]{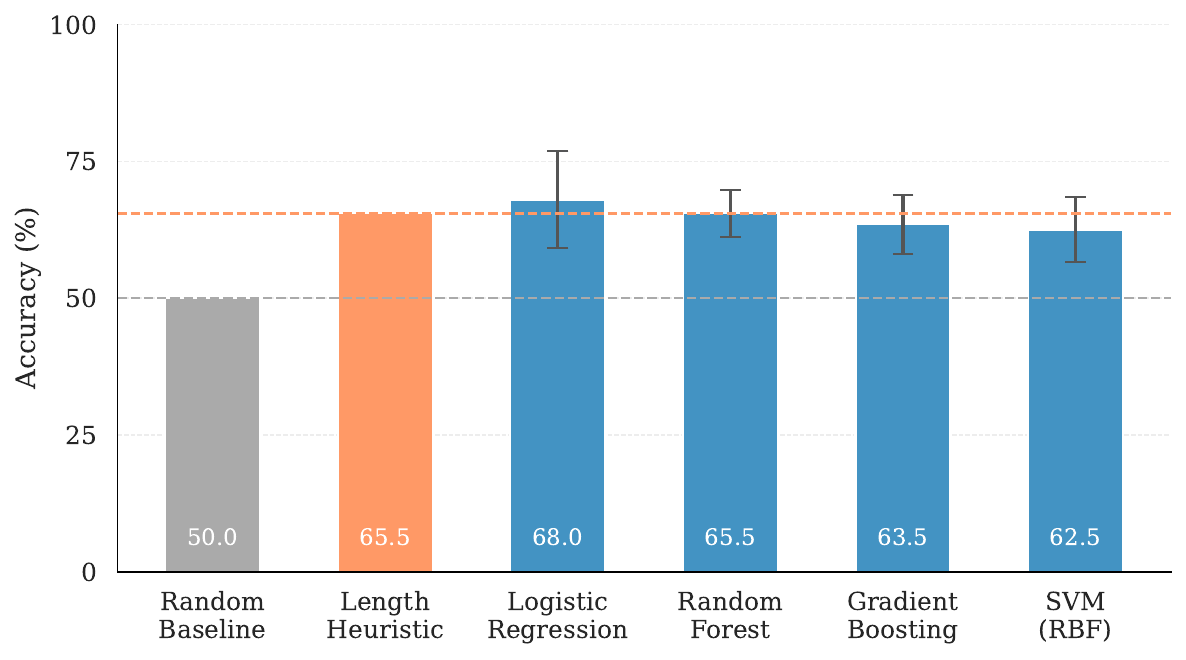}
    \caption{Non-LLM baselines on the behavioral task (5-fold CV).
    Dashed lines mark the random (50\%) and length-heuristic
    (65.5\%) baselines.}
    \label{fig:nonllm_behavioral}
\end{minipage}
\hfill
\begin{minipage}{0.39\textwidth}
    \centering
    \includegraphics[width=\linewidth]{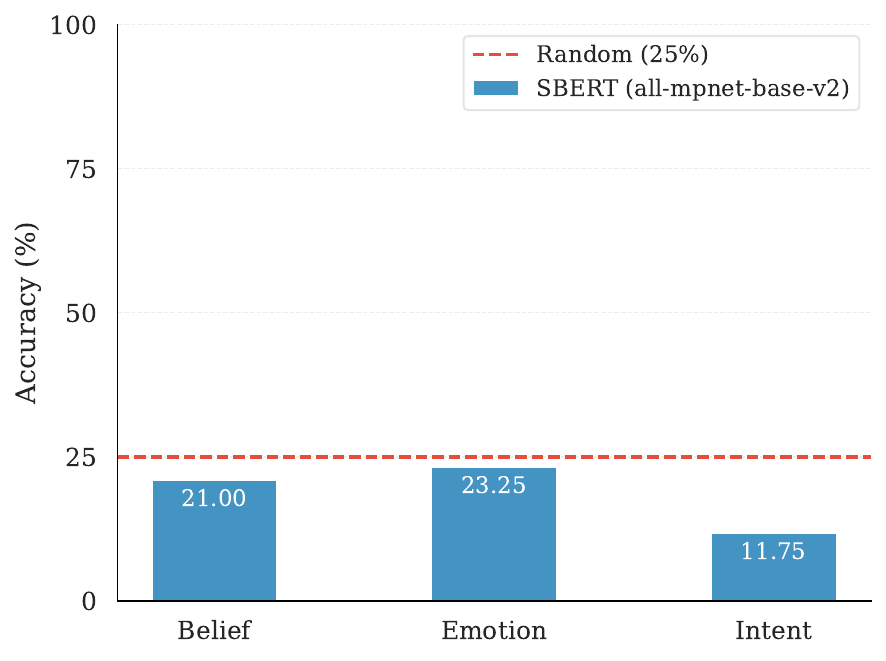}
    \caption{SBERT similarity baseline on the cognitive task. All
    dimensions fall below the 25\% chance level.}
    \label{fig:nonllm_cognitive}
\end{minipage}
\end{figure*}

\begin{figure*}[t]
\centering
\includegraphics[width=0.65\textwidth]{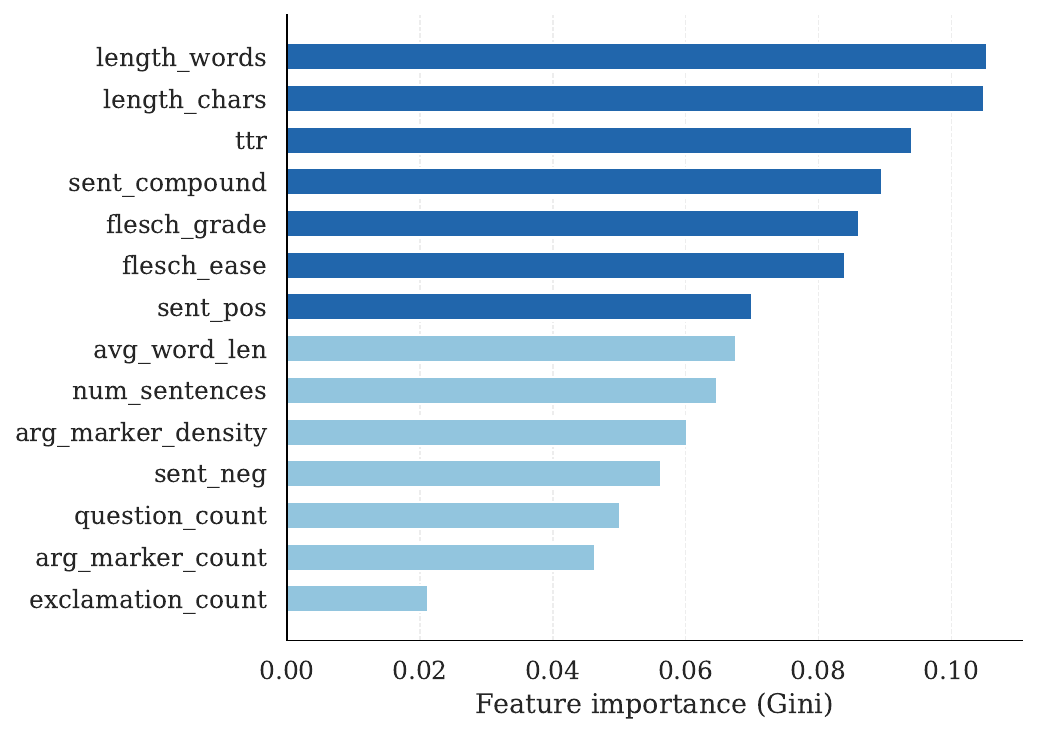}
\caption{Random Forest feature importance on the behavioral task.
Each feature is the difference between its values for the two
candidate replies.}
\label{fig:rf_importance}
\end{figure*}

Table~\ref{tab:nonllm_results} and
Figures~\ref{fig:nonllm_behavioral}--\ref{fig:rf_importance} report the
full results of the non-LLM baselines introduced in
Section~\ref{sec:evaluation_setup}.

\paragraph{Behavioral task.}
We establish three non-LLM reference points for the behavioral task:
a \textit{random} baseline (50\%), a \textit{length} heuristic that
always selects the longer reply (65.5\%), and four \textit{surface-feature
classifiers} (Logistic Regression, Random Forest, Gradient Boosting, and
SVM) trained with 5-fold cross-validation on 14 linguistic features,
including reply length, sentiment polarity, readability, lexical
diversity, and discourse marker density. These classifiers achieve
62.5--68.0\%, all within one standard deviation of the length heuristic.
The two most important Random Forest features
(Figure~\ref{fig:rf_importance}) are reply length in words and
characters, and the third, lexical diversity, is itself strongly
correlated with length ($r=-0.80$). These results indicate that the
predictive power of the examined surface features is largely driven by
reply length, suggesting that models substantially outperforming these
baselines capture signals beyond simple surface-level cues.

\paragraph{Cognitive task.}

We also evaluate a Sentence-BERT similarity baseline
(\textit{all-mpnet-base-v2})~\citep{reimers2019sentencebertsentenceembeddingsusing}
that encodes the context observed by the models (the OP's post and the
persuader's reply) and selects the most similar answer choice. It achieves 21.00\%, 23.25\%, and 11.75\% on
belief, emotion, and intent respectively, all below the 25\% chance
level, confirming that the cognitive task
cannot be resolved through surface-level semantic matching.
On intent, the gold answer is the \textit{least} similar
option in 49.2\% of instances, compared to 25\% expected by chance.
This is consistent with our generation guidelines
(Appendix~\ref{sec:appendix_Prompt}), which require gold answers to
abstract away from the literal content of the dialogue while
distractors remain bound to its context, so that surface similarity
is not a usable shortcut.

\begin{table}[h]
\centering
\small
\setlength{\tabcolsep}{8pt}
\renewcommand{\arraystretch}{1.15}
\begin{tabular}{lcc}
\toprule
\textbf{Method} & \textbf{Accuracy (\%)} & \textbf{$\pm$Std} \\
\midrule
\multicolumn{3}{l}{\textit{Behavioral task} ($n=200$)} \\[1pt]
\quad Random baseline     & 50.0 & -- \\
\quad Length heuristic    & 65.5 & -- \\
\quad Logistic Regression & 68.0 & 8.9 \\
\quad Random Forest       & 65.5 & 4.3 \\
\quad Gradient Boosting   & 63.5 & 5.4 \\
\quad SVM (RBF)           & 62.5 & 5.9 \\
\midrule
\multicolumn{3}{l}{\textit{Cognitive task, SBERT} ($n=400$ per dimension)} \\[1pt]
\quad Belief              & 21.00 & -- \\
\quad Emotion             & 23.25 & -- \\
\quad Intent              & 11.75 & -- \\
\quad Random baseline     & 25.00 & -- \\
\bottomrule
\end{tabular}
\caption{Non-LLM baseline results. Std is computed across the five
cross-validation folds; deterministic methods have none.}
\label{tab:nonllm_results}
\end{table}

\subsection{Robustness to Reply Length}
\label{sec:appendix_length_control}

As the replies in \textsc{PerSo} are naturally occurring Reddit comments,
the two candidate replies in a pair can differ in length. To verify that
the behavioral evaluation is robust to such differences, we repeat the
evaluation after removing outlier instances with an extreme length
difference between the two replies. Specifically, among instances where
the successful reply is longer, we compute the ratio between the lengths
of the two replies and discard those above the third quartile (Q3) of
this ratio, corresponding to a successful reply more than 4.97 times
longer than its counterpart. This removes 33 instances and leaves a
subset of 167 instances.

Table~\ref{tab:length_control} compares model accuracy on the full
dataset and on this subset. Removing these instances leads to a modest
decrease in accuracy for all models, while the overall pattern of
results is preserved: the relative ordering of models remains largely
unchanged, and Qwen-3.5-397B remains the best-performing model. These
results indicate that length differences between the replies have a
limited effect on the behavioral evaluation, and that the conclusions
drawn from the full dataset remain valid.
\begin{table}[h]
\centering
\small
\setlength{\tabcolsep}{6pt}
\renewcommand{\arraystretch}{1.1}
\begin{tabular}{lcc}
\toprule
\textbf{Model} & \textbf{Full (200)} & \textbf{Filtered (167)} \\
\midrule
Gemma-4B           & 60.50 & 56.29 \\
Gemma-12B          & 62.50 & 59.88 \\
Gemma-27B          & 71.00 & 67.66 \\
\midrule
Qwen-4B            & 64.00 & 57.49 \\
Qwen-35B           & 75.00 & 70.06 \\
Qwen-397B          & \textbf{76.00} & \textbf{71.26} \\
\midrule
Qwen-4B$\dagger$   & 66.50 & 61.08 \\
Qwen-35B$\dagger$  & 73.00 & 68.26 \\
Qwen-397B$\dagger$ & 73.00 & 70.06 \\
\midrule
DeepSeek-V3.1      & 67.50 & 61.68 \\
DeepSeek-V3.2      & 69.00 & 62.87 \\
DeepSeek-V4-Pro    & 62.00 & 58.08 \\
\midrule
Haiku-4.5          & 69.50 & 65.27 \\
Sonnet-4.6         & 71.50 & 67.07 \\
\midrule
Gemini-2.5-Pro     & 71.00 & 67.07 \\
\bottomrule
\end{tabular}
\caption{Behavioral accuracy (\%) on the full dataset and on the
filtered subset obtained by removing instances with an extreme length
difference between the two replies (length ratio above Q3).
$\dagger$ denotes models evaluated with thinking enabled.}
\label{tab:length_control}
\end{table}

\subsection{Prompt Templates}
\label{sec:appendix_Prompt}
We detail the prompt templates used in our LLM-in-the-loop pipeline for Cognitive MC generation and model evaluation. 
\paragraph{Cognitive MC Generation} We employ a two-step, few-shot prompting strategy governed by generation guidelines to construct the cognitive evaluation dataset. First (Figure~\ref{fig:gold_label_prompt}), the model receives the \texttt{<FULL\_DIALOGUE>} to generate the gold choice for the persuadee's belief, emotion, and intent based on their final utterance. Second (Figure~\ref{fig:distractors_prompt}), the model uses these extracted gold answers to generate three plausible, definitively incorrect distractors.
\paragraph{Cognitive Evaluation Task} To assess social reasoning, the model receives \texttt{<BODY\_UTTERANCE>} (Figure~\ref{fig:cognitive_prompt}) and must select the multiple-choice option that best predicts the persuadee's upcoming belief, emotion, or intent in the next turn.
\paragraph{Behavioral Evaluation Task} To evaluate behavioral intelligence the model receives \texttt{<OP\_INITIAL\_POST>} and two related \texttt{<PERSUADER'S\_COMMENT>} (Figure~\ref{fig:behavioral_prompt}). To mitigate position bias, the model assigns unequal scores (1--4) to rate each candidate's persuasiveness. The higher-scoring response represents the model's selection for the more effective social action.

\begin{center}
    \begin{tcolorbox}[
        breakable,
        colback=black!5, 
        colframe=black!75, 
        colbacktitle=black!75, 
        coltitle=white, 
        title={Prompt for generating gold label},
        arc=2mm, 
        boxrule=0.75pt, 
        left=2mm, right=2mm, top=2mm, bottom=2mm 
    ]
    \small 
    You are an expert at analyzing persuasion dialogues.  \\
    Your task is to extract the persuadee's true belief, intent, and emotion at the end of a dialogue.  

    For each factor (belief, intent, emotion), generate a correct answer that represents the underlying mental state of the persuadee in their last utterance.
\begin{itemize}
       \item \textbf{Belief} is an internal representation of what an agent considers true.
       
        \item \textbf{Emotion} refers to the feeling state of an agent, often conveyed by cues.
       
        \item \textbf{Intent} refers to goals, motivation, and the underlying reasons for actions.
\end{itemize}

    \textbf{Important Answer Generation Notes:}
    \begin{itemize}
        \item[-] \textbf{->>} The correct answer should reflect the cognitive aspect of the persuadee expressed at the \texttt{<<LAST ONLY UTTERANCE>>}. Avoid expressing the persuadee's cognitive aspects in utterances other than last in your correct answer.
        \item[-] \textbf{->>} Focus on the cognitive state, NOT OVERLY on the LITERAL DETAILS of the utterance --- avoid restating content (no quotes, no paraphrase of what was said), do NOT mention exact quoted phrases from the dialogue.
    \end{itemize}
    
    \textbf{For Example:} \\
    \textit{Dialogue's Last Utterance:} \\
    ``I disagree, still. While Reddit does allow for some semblance of anonymity, its content still remains public domain. Reddit gathers a wide variety of users who span a massive spectrum of what personal information they do, and do not, share. The fact that you created a user name does now deter from the IP traces an individual, or organization in this case, can do. When someone gains notoriety like the redditor in question did, it simply puts an even bigger spotlight on their account. An account which, like it or not, remains tied to the actual person. So contextually, Twitter and Reddit are inherently the same.''
    
    \textit{Bad Answer (Too Specific):} \\
    - Belief: The persuadee still believes that Reddit and Twitter are contextually the same because Reddit's content is public domain and a user's anonymity is not guaranteed, making the situations of the Redditor and Kathy Griffin comparable. \\
    - Intent: To rebut the persuader's argument by explaining why the perceived anonymity on Reddit is not a valid distinction from posting on a platform like Twitter, thus reinforcing their original stance. \\
    - Emotion: Frustrated and annoyed that the persuader is not understanding the technical aspects of internet anonymity.
    
    \textit{Good Answer (Abstracted Enough):} \\
    - Belief: Still believes the discussed platforms are similar because both operate in the public domain. \\
    - Intent: To challenge a comparison by arguing that the two platforms being contrasted are fundamentally similar. \\
    - Emotion: Resolute

    \begin{itemize}
        \item[-] \textbf{->>} As an additional checklist: Imagine the last utterance were worded differently, make sure your answer should still apply.
        \item[-] \textbf{->>} Keep your answers compact, following the style shown in the following examples. Avoid generating relatively very long answers.
    \end{itemize}

    \textbf{The fixed questions to which you should provide your correct answers are:}
    \begin{itemize}
        \item ``What is the Persuadee's Belief at the end of the last turn?''
        \item ``What is the Persuadee's Emotion at the end of the last turn?''
        \item ``What is the Persuadee's Intent at the end of the last turn?''
    \end{itemize}

    \textbf{Here's your thinking process for generating the JSON:}
    \begin{enumerate}
        \item Read the entire dialogue to understand the persuadee's initial stance and how it evolves.
        \item Analyze the final utterance as the endpoint of that evolution.
        \item For each category (Belief, Emotion, Intent), infer the persuadee's internal state.
        \item Draft the answer, ensuring it follows the Key Principles and avoids the forbidden elements.
    \end{enumerate}

    Return the output strictly as valid JSON in the following schema:
\begin{lstlisting}[basicstyle=\ttfamily\small, breaklines=true]
{
  "belief_question": {
    "question": "What is the persuadee’s Belief at the end of the last turn?",
    "answer": "[Belief's correct answer]"
  },
  "emotion_question": {
    "question": "What is the persuadee’s Emotion at the end of the last turn?",
    "answer": "[Emotion's correct answer]"
  },
  "intent_question": {
    "question": "What is the persuadee’s Intent at the end of the last turn?",
    "answer": "[Intent's correct answer]"
  }
}
\end{lstlisting}
    -->>> Do Not include any extra text outside the JSON <<<--

    \vspace{2mm}
    \textbf{Dialogue's format:}
    \begin{itemize}
        \item \textbf{A:} denotes the start of the persuadee's utterance
        \item \textbf{B:} denotes the start of the persuader's utterance
        \item {Syntax \textbf{\texttt{>statement}} marks a quote}
    \end{itemize}

    \textbf{Examples:} \\
    Case 0: \\
    ... \\
    Case 1: \\
    ... \\
    \textbf{Dialogue:} \\
    \texttt{<FULL\_DIALOGUE>}
    \end{tcolorbox}
\captionof{figure}{Prompt template for generating gold labels.} 
    \label{fig:gold_label_prompt}
\end{center}

\begin{center}
\begin{tcolorbox}[
    breakable, 
    colback=black!5, 
    colframe=black!75, 
    colbacktitle=black!75, 
    coltitle=white, 
    title={Prompt for generating distractors},
    arc=2mm, 
    boxrule=0.75pt, 
    left=2mm, right=2mm, top=2mm, bottom=2mm 
]
\small 

You are an expert at analyzing persuasion dialogues. \\
Your task is to generate distractors for the persuadee’s belief, intent, and emotion at the end of a dialogue, given the persuadee's actual states. 

For each factor (belief, intent, emotion), generate 3 plausible distractors, bound by the dialogue’s context, given the persuadee’s actual states. Ensure that the distractors match the correct answer in their level of abstraction.

\begin{itemize}
        \item \textbf{Belief} is an internal representation of what an agent considers true.
        \item \textbf{Emotion} refers to the feeling state of an agent, often conveyed by cues.
        \item \textbf{Intent} refers to goals, motivation, and the underlying reasons for actions.
    \end{itemize}

\textbf{Distractor generation notes:}
\begin{itemize}
    \item[-] \textbf{->} Persuadee's actual states at the last utterance (the correct answers for which you should make distractors) are provided at the end of each dialogue.
    \item[-] \textbf{->} Avoid Shortcut Effects when creating distractors. Shortcut effect refers to situations where distractors are easy to eliminate without analyzing the context provided. Here's an example of the shortcut effect: \\
    Q: What is the Persuadee’s emotion at the end of the last turn? \\
    A (Distractor) : Reluctant and hesitant, as they feel forced to admit they were wrong. \\
    A (Distractor with Shortcut) : In love, and joyful, as they have found a deep connection with the persuader. \\
    It is obvious that in a persuasion scenario, the case of the persuadee falling in love with the persuader is hardly possible, therefore the second choice proposes a shortcut effect, where no context analysis is needed for eliminating this choice.
    \item[-] \textbf{->} Avoid creating distractors that are ``technically'' correct. Designed distractors must be definitely wrong.
    \item[-] \textbf{->} Avoid creating distractors that imply mean behavior, for example: \\
    Q: What is the Persuadee’s intent at the end of the last turn, reflected at the last utterance? \\
    A (Distractor) : To request more information about the points raised in the persuader's second paragraph. \\
    A (Distractor implying mean behavior) : To start an outburst of profanity, trying to offend the persuader’s beloved ones.
    \item[-] \textbf{->} Make sure not to create distractors that are too plausible or can be correct in general, or are overly customized and narrowly focused on dialogue's last utterance.
    \item[-] \textbf{->} Make distractors that are incorrect in the way of not following persuadee's mental state's rational flow in the dialogue.
    \item[-] \textbf{->} Make sure all distractors and the correct answer are distinct and all the distractors are definitely wrong.
    \item[-] \textbf{->} Make sure NOT to mention exact phrases from the dialogue in the distractors.
    \item[-] \textbf{->} Make sure to preserve the style of the correct answer (e.g., phrasing, tone, and length) when writing distractors. For emotion questions specifically, emphasize that the length should be a short phrase, like the correct answer.
\end{itemize}

\textbf{The fixed questions for which you should make distractors are:}
    \begin{itemize}
        \item ``What is the Persuadee’s Belief at the end of the last turn?''
        \item ``What is the Persuadee’s Emotion at the end of the last turn?''
        \item ``What is the Persuadee’s Intent at the end of the last turn?''
    \end{itemize}

Return the output strictly as valid JSON in the following schema:
\begin{lstlisting}[basicstyle=\ttfamily\small, breaklines=true]
{{
  "belief_question": {{
    "question": "What is the persuadee’s Belief at the end of the last turn?",
    "distractors":  [
        "1) [Distractor text]"
        "2) [Distractor text]"
        "3) [Distractor text]"
    ]
  }},
  "emotion_question": {{
    "question": "What is the persuadee’s Emotion at the end of the last turn?",
    "distractors":  [
        "1) [Distractor text]"
        "2) [Distractor text]"
        "3) [Distractor text]"
    ]
  }},
  "intent_question": {{
    "question": "What is the persuadee’s Intent at the end of the last turn?",
    "distractors":  [
        "1) [Distractor text]"
        "2) [Distractor text]"
        "3) [Distractor text]"
    ]
  }}
}}
\end{lstlisting}
-->>> Do Not to include any extra text outside the JSON <<<--

\vspace{2mm}
\textbf{Dialogue's format:}
\begin{itemize}
\item \textbf{A:} denotes the start of the persuadee's utterance
\item \textbf{B:} denotes the start of the persuader's utterance
\item Syntax \textbf{\texttt{>statement}} marks a quote
\end{itemize}

\textbf{Examples:} \\
    Case 0: \\
    ... \\
    Case 1: \\
    ... \\
    
\textbf{Dialogue:} \\
    \texttt{<FULL\_DIALOGUE>}\\
    
\textbf{Persuadee’s Actual states (Make distractors for these):} \\
\begin{itemize}
    \item Persuadee's belief:\\ \texttt{<BELIEF\_CORRECT\_ANSWER>}
    \item Persuadee's emotion:\\ \texttt{<EMOTION\_CORRECT\_ANSWER>}
    \item Persuadee's intent:\\         \texttt{<INTENT\_CORRECT\_ANSWER>}
\end{itemize}
\end{tcolorbox}
\captionof{figure}{Prompt template for generating distractors.}
    \label{fig:distractors_prompt}
\end{center}

\begin{center}
\begin{tcolorbox}[
    breakable, 
    colback=black!5, 
    colframe=black!75, 
    colbacktitle=black!75, 
    coltitle=white, 
    title={Prompt for cognitive evaluation},
    arc=2mm, 
    boxrule=0.75pt, 
    left=2mm, right=2mm, top=2mm, bottom=2mm 
]
\small 

You are an expert at analyzing persuasion dialogues. \\
You are given a persuasion dialogue between a persuadee and a persuader. \\
Your task is to answer a multiple-choice question about the persuadee’s \texttt{\{MENTAL\_STATE\_DIMENSION\}} at the end of the dialogue.

\vspace{2mm}
\textbf{Important}: \\
The provided dialogue is not yet complete. The persuadee will speak again after the last shown turn. Your job is to predict the persuadee’s \texttt{\{MENTAL\_STATE\_DIMENSION\}} in their upcoming next utterance, not to infer their past or current state.  

-->> Select the option that best reflects the persuadee’s \texttt{\{MENTAL\_STATE\_DIMENSION\}} at the end of the dialogue.

\vspace{2mm}
\textbf{Dialogue's format:}
\begin{itemize}
    \item \textbf{A:} denotes the start of the persuadee's utterance
    \item \textbf{B:} denotes the start of the persuader's utterance
    \item Syntax \textbf{\texttt{> statement}} marks a quote
\end{itemize}

\vspace{2mm}
\textbf{Dialogue:} \\
\texttt{<BODY\_UTTERANCE>}

\vspace{2mm}
\textbf{Question:} \\
What will the persuadee’s \texttt{<MENTAL\_STATE\_DIMENSION>} be in their next upcoming utterance after the end of the shown dialogue?

\vspace{2mm}
\textbf{Choices:}
\begin{itemize}
    \item \texttt{<GENERATED\_CHOICE\_1>}
    \item \texttt{<GENERATED\_CHOICE\_2>}
    \item \texttt{<GENERATED\_CHOICE\_3>}
    \item \texttt{<GENERATED\_CHOICE\_4>}
\end{itemize}

-->>> Provide ONLY THE CORRECT ANSWER LETTER (A, B, ...) in the response, NO EXTRA TEXT <<<--
\end{tcolorbox}
\captionof{figure}{Prompt template for evaluating cognitive intelligence.}
    \label{fig:cognitive_prompt}
\end{center}

\begin{center}
\begin{tcolorbox}[
    breakable, 
    colback=black!5, 
    colframe=black!75, 
    colbacktitle=black!75, 
    coltitle=white, 
    title={Prompt for rating persuasive utterances},
    arc=2mm, 
    boxrule=0.75pt, 
    left=2mm, right=2mm, top=2mm, bottom=2mm 
]
\small 

You are an expert in persuasion and communication strategies.

\vspace{2mm}
\textbf{You are given:}
\begin{itemize}
    \item A description of a persuadee's current view.
    \item Two candidate persuasive utterances (A and B).
\end{itemize}

Rate each utterance.

\vspace{2mm}
\textbf{Scoring:}
\begin{itemize}
    \item \textbf{1} = Not persuasive
    \item \textbf{2} = Slightly persuasive
    \item \textbf{3} = Persuasive
    \item \textbf{4} = Very persuasive
\end{itemize}

\textbf{Important Guidelines:}
\begin{itemize}
    \item The two scores \textbf{MUST NOT} be equal.
    \item If they seem equally persuasive, choose the one that is even slightly better.
\end{itemize}

\vspace{2mm}
-->>> \textbf{Output ONLY in this format:} <<<-- \\
\texttt{A=?} \\
\texttt{B=?}

\vspace{2mm}
\textbf{Do not output any explanation.}

\vspace{2mm}
\textbf{Note:} Syntax \textbf{\texttt{>statement}} denotes a quotation.

\vspace{2mm}
-->> \textbf{Persuadee's view:} \\
\texttt{<OP\_INITIAL\_POST>}

\vspace{2mm}
-->> \textbf{Candidate Utterances:}
\begin{itemize}
    \item \texttt{<FIRST\_PERSUADER'S\_COMMENT>}
    \item \texttt{<SECOND\_PERSUADER'S-COMMENT>}
\end{itemize}
\end{tcolorbox}
\captionof{figure}{Prompt template for evaluating behavioral intelligence.}
    \label{fig:behavioral_prompt}
\end{center}

\subsection{Processed Example from PERSO}
\label{app:sample_branch77}

\noindent
This appendix presents a processed example illustrating how a paired persuasion branch is converted into the cognitive and behavioral tasks.

\begin{figure*}[t]
\centering
\includegraphics[width=0.65\textwidth]{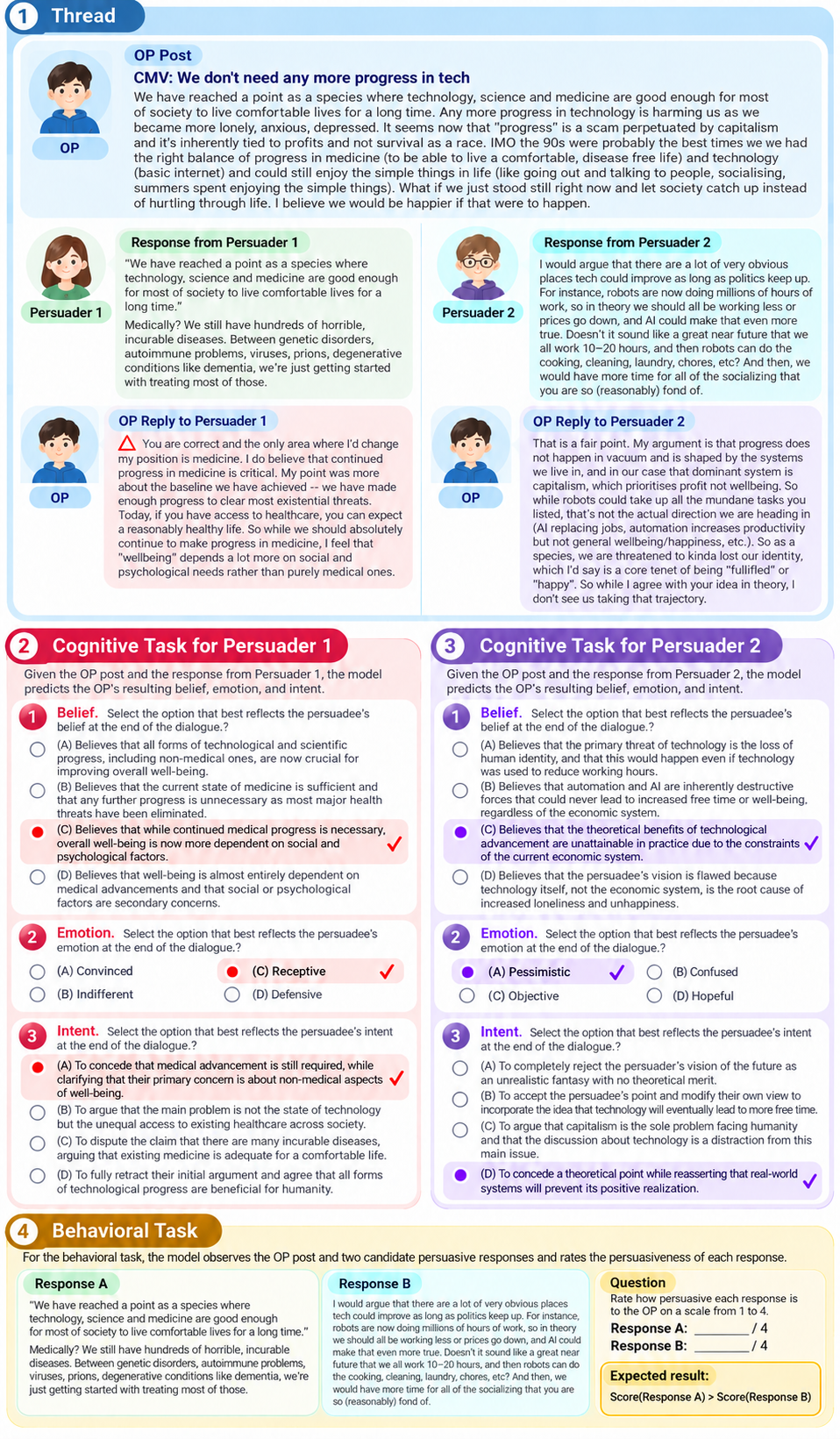}
\caption{A processed example illustrating how a paired persuasion branch is converted into the cognitive and behavioral tasks.}
\label{fig:processed_thread}
\end{figure*}

Figure~\ref{fig:processed_thread} shows the complete source thread in part~1, the cognitive task for Persuader~1 in part~2,  the cognitive task for Persuader~2 in part~3, and the corresponding behavioral comparison between the two persuader responses in part~4.

\clearpage

\end{document}